\documentclass[acmtomm,nonacm]{acmart}

\AtBeginDocument{%
  }

\usepackage{booktabs} 
\usepackage{bm}       

\usepackage[ruled,vlined,linesnumbered]{algorithm2e} 
\SetCommentSty{footnotesize\color{gray}\ttfamily} 

\begin{document}

\title{GAT-NeRF: Geometry-Aware-Transformer Enhanced Neural Radiance Fields for High-Fidelity 4D Facial Avatars}

\author{Zhe Chang}
\email{233370890@st.usst.edu.cn}
\affiliation{%
  \institution{Department of Control Science and Engineering, University of Shanghai for Science and Technology}
  \city{Shanghai}
  \postcode{200093}
  \country{China}
}

\author{Haodong Jin}
\email{231260086@st.usst.edu.cn}
\affiliation{%
  \institution{Department of Control Science and Engineering, University of Shanghai for Science and Technology}
  \city{Shanghai}
  \postcode{200093}
  \country{China}
}

\author{Ying Sun}
\email{yingsun1991@163.com}
\affiliation{%
  \institution{Business School, University of Shanghai for Science and Technology}
  \city{Shanghai}
  \postcode{200093} 
  \country{China}
}

\author{Yan Song}
\email{sonya@usst.edu.cn}
\affiliation{%
  \institution{Department of Control Science and Engineering, University of Shanghai for Science and Technology}
  \city{Shanghai}
  \postcode{200093}
  \country{China}
}

\author{Hui Yu}
\authornote{Corresponding author.}
\email{hui.yu@glasgow.ac.uk}
\affiliation{%
  \institution{School of Psychology and Neuroscience, University of Glasgow}
  \city{Glasgow}
  \country{UK}
}

\renewcommand{\shortauthors}{Z. Chang et al.}


\begin{abstract}
High-fidelity 4D dynamic facial avatar reconstruction from monocular video is a critical yet challenging task, driven by increasing demands for immersive virtual human applications. While Neural Radiance Fields (NeRF) have advanced scene representation, their capacity to capture high-frequency facial details, such as dynamic wrinkles and subtle textures from information-constrained monocular streams, requires significant enhancement. To tackle this challenge, we propose a novel hybrid neural radiance field framework, called Geometry-Aware-Transformer Enhanced NeRF (GAT-NeRF) for high-fidelity and controllable 4D facial avatar reconstruction, which integrates the Transformer mechanism into the NeRF pipeline. GAT-NeRF synergistically combines a coordinate-aligned Multilayer Perceptron (MLP) with a lightweight Transformer module, termed as Geometry-Aware-Transformer (GAT) due to its processing of multi-modal inputs containing explicit geometric priors. The GAT module is enabled by fusing multi-modal input features, including 3D spatial coordinates, 3D Morphable Model (3DMM) expression parameters, and learnable latent codes to effectively learn and enhance feature representations pertinent to fine-grained geometry. The Transformer's effective feature learning capabilities are leveraged to significantly augment the modeling of complex local facial patterns like dynamic wrinkles and acne scars. Comprehensive experiments unequivocally demonstrate GAT-NeRF's state-of-the-art performance in visual fidelity and high-frequency detail recovery, forging new pathways for creating realistic dynamic digital humans for multimedia applications.
\end{abstract}






\maketitle



% ...

% ...

\section{Introduction}
The creation of high-fidelity, dynamic four-dimensional (4D) digital humans is a central research imperative in computer vision, graphics, and multimedia, driven by their vast application potential in areas ranging from virtual production and immersive AR/VR to personalized healthcare and forensic science~\cite{Sun24MultiModalDriven,Zou24_4DFacialExpression, Liu24MultimodalFusion,Liu23TalkingFaceAnatomy,Pesavento24ANIM,cao2023high}. In the healthcare domain, for example, detailed facial features analysis is critical for computer-aided diagnosis of conditions such as facial palsy, highlighting the need for robust facial data and models~\cite{xia2022aflfp}. A significant technical challenge persists in reconstructing animatable head avatars from accessible monocular videos that can meticulously reproduce high-frequency geometric and textural details, such as dynamic wrinkles and personalized scars. This research endeavors to address this challenge by proposing a monocular video-driven framework for high-fidelity, detail-rich dynamic head avatar reconstruction.

Neural Radiance Fields (NeRF)~\cite{Mildenhall21Nerf} have emerged as a seminal paradigm for 3D reconstruction and novel view synthesis, learning a continuous scene representation from images. Its variants have been successfully applied to model diverse scenes, including human faces~\cite{Athar22Rignerf,Grassal22NeuralHead, Bergman22GenerativeNeural,Cui24AlethNerf,Zhan24KFDNeRF,Jiang22Selfrecon,Zheng22ImAvatar}. While explicit methods like 3D Gaussian Splatting (3DGS)~\cite{Kerbl23GaussianSplatting,Fei24GaussianSplattingSurvey,Ali25CompressionGaussian,Bao25_3DGaussianSplattingSurvey} offer remarkable rendering speeds, the implicit and continuous nature of NeRF provides a malleable foundation for our goal of reconstructing highly detailed, dynamically controllable facial avatars. NeRF's per-point querying mechanism is particularly advantageous for integrating potent feature learning modules and fusing spatial information with geometric priors from models like the 3D Morphable Model (3DMM) and learned latent codes, guiding precise attribute prediction.

Despite NeRF's success in capturing macroscopic geometry~\cite{Athar22Rignerf,Zielonka23InstantVolumetric,Siarohin19FirstOrderMotion,Hong22Headnerf,Gafni21DynamicNeural,Zheng23Pointavatar,Bergman22GenerativeNeural,Jiang22Selfrecon}, accurately reconstructing facial micro-structures—especially dynamic high-frequency details like expression-induced wrinkles or subtle identity-specific markings—remains a challenge. Traditional MLP-based NeRF architectures, with their limited receptive fields, often struggle to represent the complex geometric dependencies underlying these fine details, leading to artifacts like blurring or over-smoothing~\cite{Zielonka23InstantVolumetric,Siarohin19FirstOrderMotion,Hong22Headnerf}. Monocular setups further exacerbate these issues due to depth ambiguity and feature degradation, demanding more robust neural representations for achieving geometric consistency~\cite{Gafni21DynamicNeural,Zheng23Pointavatar}.

To address these limitations, we propose GAT-NeRF in this study: a Transformer-enhanced Neural Radiance Field framework. This hybrid architecture synergistically integrates a coordinate-aligned MLP with a lightweight Transformer module, termed as Geometry-Aware Transformer (GAT) due to its designated role in processing the constructed multi-modal inputs containing explicit geometric priors. The GAT-NeRF framework's core design enables a standard Transformer encoder to focus on learning and enhancing features pertinent to fine-grained facial geometry by providing it with fused input features comprising: (i) positional encodings $\text{PE}(\mathbf{p})$ of 3D spatial coordinates, (ii) expression parameters $\bm{\delta}$ from a 3DMM, and (iii) learnable latent codes $\bm{\gamma}$ for personalization and error compensation. The Transformer's inherent feature learning capabilities are thus effectively leveraged to augment the model's proficiency in representing complex local patterns associated with high-frequency details (e.g., dynamic wrinkles). This synergistic operation of the GAT module with the subsequent MLP facilitates efficient multi-modal feature fusion and directly guides the MLP towards more accurate color and density prediction, achieving dynamic facial reconstruction with excellent geometric consistency and detail.

The proposed GAT-NeRF hybrid architecture establishes an effective new methodology for high-fidelity facial reconstruction driven by a monocular video, significantly advancing the capability for preserving and generating high-frequency dynamic details. The primary contributions of this paper are summarized as follows:
\begin{enumerate}
\item We introduce GAT-NeRF, a novel hybrid neural radiance field framework that efficiently integrates the Transformer mechanism with NeRF. By fusing spatial coordinates with multi-modal inputs, 3DMM expression parameters, and personalized latent codes, GAT is enabled to effectively learn and enhance the representation of fine facial geometric details.
\item By integrating the Transformer mechanism within the NeRF framework and leveraging its capacity to process specific geometry-related inputs, our method significantly optimizes the reconstruction quality of high-frequency facial details, enabling the accurate recovery and preservation of subtle textures and dynamic geometric features corresponding to evolving facial expressions.
\item Extensive experiments demonstrate the excellent performance of our method in terms of fidelity and visual realism in reconstructing high-frequency facial details under monocular video input.
\end{enumerate}

\section{Related Work}
\subsection{Facial Avatar Reconstruction, Synthesis, and Control}
Advances in deep learning have spurred innovations in numerous domains, such as natural language processing~\cite{Shamshiri24TextMining,Wang23ChatGPTDAO,Ahmed24LinguisticIntelligence}, autonomous driving~\cite{Zhao24AutonomousDriving,Yu23SocialVision}, and intelligent transportation~\cite{Zhou24AdaptiveSegmentation,Bakirci24EnhancingVehicle,Chen20CitywideTraffic}.In parallel, significant progress has been made in the understanding of high-level video, for example, by generating textual descriptions for micro-videos~\cite{Ge25Implicit, Qiao25Multimodal} or leveraging multi-modal cues like sound to interpret video content~\cite{Liu24Enhancing, Jing24Multimodal}. These advancements highlight the growing capabilities of models to process and interpret dynamic visual data, which motivates the development of more sophisticated video-driven applications. Among these, Reconstructing high-fidelity digital human faces from monocular video has attracted significant attention due to its vast application potential. Traditionally, the pioneering work of Blanz and Vetter~\cite{Blanz23MorphableModel} introduced 3DMM, which modeled face shape and appearance in a low-dimensional linear subspace via Principal Component Analysis (PCA). Many subsequent head reconstruction methods~\cite{Thies19DeferredNeural,Buehler21Varitex,Yu12PerceptionDriven,Xia21RelationAwareB} leveraged parameterized 3DMMs to achieve reasonable reconstruction results. However, their reliance on a fixed template makes it challenging to recover non-parametric regions like hair and accessories, and they face limitations in expressing fine-grained, personalized facial details.

On another front, 2D face synthesis methods based on Generative Adversarial Networks (GANs)~\cite{Luo22DualGGAN,Yu22CMOSGAN,Kammoun22GANSurvey,Chibane21StereoRadiance,Choi18Stargan,Kwak20CafeGAN} can produce photorealistic images. Nevertheless, these 2D GAN models inherently lack 3D spatial awareness, leading to fundamental challenges in maintaining multi-view consistency. Recent progress in 3D-aware GANs~\cite{Deng22Gram,Chan22EfficientGeometry} has partially mitigated spatial consistency issues but still falls short in modeling complex facial dynamics and fine geometry. More recently, 3DGS~\cite{Kerbl23GaussianSplatting} has significantly accelerated 3D reconstruction by introducing explicit 3D Gaussian primitives and developing an efficient rasterization-based rendering algorithm. Subsequent works~\cite{Xiang24Flashavatar,Wu24_4DGaussianSplatting,Yang24DeformableGaussians,Chen24TextTo3DGaussian,Tang23Dreamgaussian,Yi24GaussianDreamer} extended it to dynamic scenes by constructing temporally-dependent deformation fields to drive the Gaussians. The core advantage of 3DGS lies in the excellent rendering speed caused by its explicit representation. However, when the research task focuses on the high-fidelity reconstruction and precise control of dynamic faces with complex micro-textures, we believe that NeRF's implicit continuous representation possesses unique advantages. As stated in the latest survey by Bao et al.~\cite{Bao25_3DGaussianSplattingSurvey}, extracting high-quality surface representations from Gaussian primitives remains challenging, which aligns with our task's requirement for extreme surface smoothness. Crucially, the survey emphasizes when discussing head reconstruction that capturing complex facial expressions is a major difficulty, and it usually requires introducing advanced MLP-based modules to enhance the expressive power of Gaussian primitives. This observation is highly consistent with our motivation, as it indicates that, for modeling fine-grained dynamic details, researchers also often resort to continuous function approximators, and NeRF provides a more native framework for such tasks.

Admittedly, recent pioneering works such as SplatTalk~\cite{Thai25SplatTalk}, 3D Vision-Language Gaussian Splatting~\cite{Peng243DVLGS}, and AnySplat~\cite{Jiang25AnySplat} have successfully integrated 3DGS with deep learning modules. But we note that their research objectives are different from our direction: these works mainly focus on the fusion of semantic information or improving the quality of semantic representations, as well as enhancing the model's cross-scene generalization ability. In contrast, our work focuses on a different challenge: achieving photo-level realism for a specific individual and precisely modeling how micro-geometry continuously evolves with expressions. Therefore, we choose NeRF because its point-wise query coordinate network architecture provides a more natural learning framework for our designed GAT module.

\subsection{Neural Radiance Fields and Their Application to Dynamic Faces}
Diverging from traditional explicit modeling of geometry and appearance, NeRF~\cite{Mildenhall21Nerf} introduced an innovative implicit representation based on Multilayer Perceptrons (MLPs). This paradigm maps a continuous 5D function (3D spatial coordinates + 2D viewing direction) to the color and volume density. Combined with advanced volume rendering techniques, NeRF enables photorealistic and view-consistent image synthesis, opening new avenues for facial avatar reconstruction.

However, classical NeRF is primarily designed for static scenes and faces inherent limitations when directly applied to dynamic scenarios~\cite{Mazzacca23NeRFHeritage,Zhang20NeRFplusplus,Lin20SDFSRN,Kato20DifferentiableRenderingSurvey}. To address spatio-temporal constraints, researchers have proposed various dynamic NeRF variants~\cite{Pumarola21DNerf,Ost21NeuralSceneGraphs,Tretschk21NonRigidNeural,Pumarola21DNerf,Gafni21DynamicNeural,Hong22Headnerf,Zheng23Pointavatar}. For instance, some methods incorporate time as an additional input or learn deformation fields to map points from a canonical space to the observed space, thereby capturing the temporal evolution of the scene~\cite{Pumarola21DNerf}. In the facial domain, Gafni et al.~\cite{Gafni21DynamicNeural} combined NeRF with parametric deformation models (e.g., 3DMMs) to achieve explicit semantic control over expression and pose while retaining the rendering advantages of radiance fields. HeadNeRF~\cite{Hong22Headnerf} modulates NeRF with expression codes for real-time facial expression reconstruction and rendering. PointAvatar~\cite{Zheng23Pointavatar} proposed a point-based deformable representation to overcome the limitations of parametric models. While these methods have made significant strides in dynamic facial modeling, their reliance on MLP-based architectures still presents room for improvement in the expressive power needed to capture and reconstruct extremely fine-grained, high-frequency facial details, such as dynamic wrinkles and personalized skin textures. This motivates our exploration of more potent feature learning mechanisms. Recently, FlashAvatar~\cite{Xiang24Flashavatar}, based on 3DGS, was the first to introduce 3DGS techniques to dynamic face reconstruction, significantly reducing training and rendering times, yet it still shows limitations in recovering high-frequency information for highly realistic faces.

It is worth noting that some works have sought to enhance the detail-capturing capability of MLP-based decoders without altering the MLP's core structure. For instance, Pixel Codec Avatars (PiCA)~\cite{Ma21Pixel} introduced a compelling approach to high-fidelity avatar rendering, especially on resource-constrained devices. To capture high-frequency details like skin pores and wrinkles, PiCA employs learned non-parametric positional encoding maps. Instead of using fixed sinusoidal functions, it learns high-resolution feature maps that store detailed appearance information at specific UV coordinates. A lightweight MLP then queries these maps to render the final pixel color. This memory-based strategy effectively offloads the complexity of detail representation to the learned maps, enabling a very efficient decoder. While highly effective, this approach relies on large, pre-learned 2D texture-space maps for detail retrieval. In our work, we explore a different, computation-driven philosophy. Rather than relying on a memory-based lookup mechanism, our goal is to enhance the intrinsic feature learning power of the network itself. We propose to integrate GAT directly into the NeRF pipeline. The GAT module performs dynamic, on-the-fly feature enhancement on multi-modal inputs, leveraging the Transformer's attention mechanism to model the complex local dependencies underlying fine-grained facial details. This allows us to directly address the representational limitations of standard MLPs, forging a new path for high-fidelity reconstruction.

\subsection{Transformer-Augmented Neural Radiance Fields}
Transformer~\cite{Vaswani17Attention}, initially designed for sequence modeling in Natural Language Processing (NLP), have garnered widespread attention in various fields due to their powerful modeling capabilities~\cite{Han22VisionTransformerSurvey,Peng23RWKV,Friedman23LearningTransformerPrograms,Hoover23EnergyTransformer,Amatriain23TransformerModelsIntro,Han23FlattenTransformer,Yao23DualVisionTransformer,Gao23GeneralizedRelationModeling}. In facial animation and reconstruction, researchers have begun to explore leveraging Transformers to enhance model performance. For example, Fan et al.~\cite{Fan22Faceformer} utilized self-attention mechanisms to model the temporal dependencies of facial movements for more realistic and smooth 3D animation. Such advanced modeling is essential for capturing the nuances of genuine emotional expressions, which are inherently complex and can be cross-validated with other modalities such as physiological signals~\cite{wang2023mgeed}. Chen et al.~\cite{Chen22Transformer3DFace} employed Transformers to capture geometric structures and texture details, improving the reconstruction accuracy. Zhang et al.~\cite{zhang2022transnerf} proposed TransNeRF, which uses a Transformer's cross-attention mechanism to condition NeRF on features extracted from multiple input views. However, Transformer-centric networks typically demand large-scale datasets and substantial computational resources, posing challenges for scenarios relying on readily accessible monocular face videos and training/inference on consumer-grade GPUs.

Some studies on general NeRF, such as the pioneering IBRNet~\cite{Wang21Ibrnet},GeoNeRF~\cite{Johari22Geonerf} and MVSNeRF~\cite{Chen21Mvsnerf}, have shown great potential in using attention or attention-like mechanisms to solve the cross-scene generalization problem. A common feature of these methods is to use Transformer or similar structures as cross-view feature aggregators to solve the complex challenges of multi-view matching and robust feature fusion in new scenes. For example, IBRNet~\cite{Wang21Ibrnet} introduced the ray transformer to perform self-attention between sample points along the ray for visibility reasoning, GeoNeRF~\cite{Johari22Geonerf} used a multi-head attention mechanism to fuse features transformed according to geometric priors. Although MVSNeRF~\cite{Chen21Mvsnerf} does not use an explicit Transformer, it also borrows the idea of multi-view stereo matching (MVS) to implicitly encode and fuse cross-view information by constructing a cost volume.

Inspired by these advancements and noting the limitations of existing dynamic NeRFs in fine-detail representation, we sought a relatively lightweight solution to effectively enhance the feature expressive power of NeRF. We leverage the unique architectural paradigm of Transformer layers—including multi-head self-attention, feedforward networks (FFN), and residual connections—by integrating it as a feature enhancement module with the NeRF. Specifically, in our GAT-NeRF framework, we design a lightweight GAT module. This module takes as input the 3D spatial coordinates, 3DMM-derived expression parameters, and learned personalized latent codes for each sampled point. By performing deep non-linear transformations and dynamic feature re-weighting on these concatenated, multi-modal, geometry-rich features, the GAT module significantly enhances the ability to discern and represent complex local patterns associated with high-frequency details. This per-point enhancement approach avoids the substantial computational overhead of global self-attention mechanisms often found in Transformers operating on point clouds or image patches, while effectively harnessing the powerful feature learning capabilities of Transformers. This allows for the more precise preservation of complex skin micro-structures, including forehead wrinkles and acne scars, ultimately improving the fidelity of local detail reconstruction.

Although our method borrows the idea of an attention mechanism at a macro-level, it is fundamentally different from~\cite{Wang21Ibrnet, Johari22Geonerf, Chen21Mvsnerf} in terms of goals and mechanisms. GAT-NeRF is designed for subject-specific high-fidelity reconstruction rather than generalization across scenes. Therefore, instead of processing image features from multiple source views, our GAT module performs point-wise feature enhancement on a fused vector containing geometric priors (e.g., spatial coordinates, 3DMM expression parameters) and learnable latent codes.  Furthermore, the attention mechanisms in~\cite{Wang21Ibrnet, Johari22Geonerf} primarily address multi-view matching and occlusion issues, whereas our GAT module is designed to solve a complex regression problem: modeling how explicit expression parameters map to microscopic, high-frequency surface details (e.g., dynamic wrinkles). This focus on leveraging geometric priors to enhance the expressive power of intra-object details constitutes the core innovation of our application of the attention mechanism, which stands in stark contrast to the idea of aggregating inter-scene information in generalizable NeRF frameworks.

\begin{figure}[htb]
  \centering
  \includegraphics[width=\columnwidth]{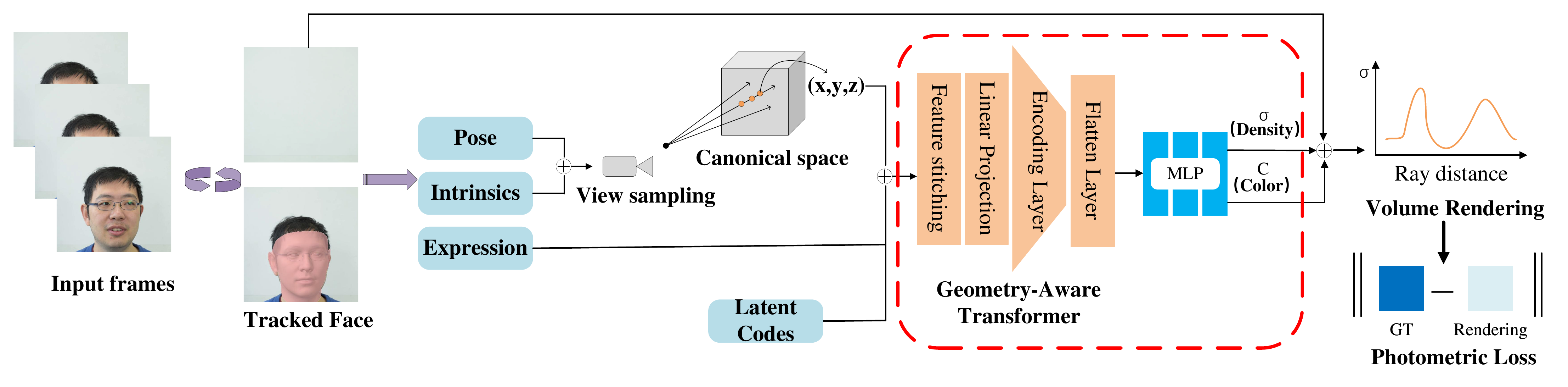}
  \Description{Overview of the proposed GAT-NeRF framework. For each input frame, 3D coordinates (p) are sampled along camera rays in a canonical space; these coordinates, facial expression parameters (δ) from a 3DMM, and a learned latent code (γ) are positionally encoded and concatenated. A Geometry-Aware Transformer (GAT) module then applies self-attention and non-linear transforms to dynamically reweight and enhance the per-point features. Finally, an MLP predicts view-dependent color (c) and density (σ) for volume rendering, capturing high-frequency facial details (e.g., dynamic wrinkles and acne scars).}
  \caption{Overview of the proposed GAT-NeRF framework. For each input frame, 3D coordinates ($\mathbf{p}$) are sampled along camera rays in a canonical space. These coordinates, along with corresponding facial expression parameters ($\bm{\delta}$) derived from a 3DMM and a learnable latent code ($\bm{\gamma}$), are positionally encoded and concatenated. This fused multi-modal feature vector is then processed by our Geometry-Aware Transformer (GAT) module. The GAT leverages its self-attention mechanism and non-linear transformations to perform dynamic feature re-weighting and enhancement on these per-point, geometry-rich inputs. The resulting enhanced features then guide a subsequent MLP to predict view-dependent color ($\mathbf{c}$) and density ($\sigma$), enabling the detailed capture of high-frequency facial attributes (e.g., dynamic wrinkles and acne scars) via volume rendering.}
  \label{fig:overview}
\end{figure}

\section{Methodology}
Our primary objective is to develop a robust framework for reconstructing high-fidelity, dynamically controllable 4D facial avatars from monocular video sequences. To achieve this, we introduce GAT-NeRF, a neural representation that synergistically integrates a novel GAT module within the NeRF paradigm. This section provides a comprehensive exposition of the GAT-NeRF framework, detailing its overall architecture, the design intricacies of the core GAT module, the mechanisms for density and color prediction, and the optimization objectives employed during training.

\subsection{Overall Architecture of GAT-NeRF}
The proposed GAT-NeRF framework employs a GAT, operating within a canonical representation space, in conjunction with a MLP network to model the dynamic radiance field of a human head, particularly during speech or expression changes. This radiance field is formulated as a function that dynamically adapts based on 3D spatial coordinates $\mathbf{p}$, 2D viewing directions $\vec{v}$, and per-frame facial expression parameters $\delta$. The comprehensive model can be mathematically expressed as:

\begin{equation}
    \mathcal{D}_{\theta}\left( 
        \mathcal{T}_{\phi}\left( 
            \text{PE}(p), \delta, \gamma 
        \right), 
        \text{PE}(\overrightarrow{v}) 
    \right) = (\text{c}, \sigma)
    \label{eq:model}
\end{equation}
where $\text{PE}(\cdot)$ denotes the high-frequency positional encoding function applied to spatial coordinates $\mathbf{p}$ (10 frequency bands) and viewing directions $\vec{v}$ (4 frequency bands), following the methodology of Mildenhall et al.~\cite{Mildenhall21Nerf}. The term $\mathcal{T}_{\phi}$ represents our GAT module, parameterized by $\phi$, which processes the concatenated positional encoding of coordinates, expression parameters $\delta$, and a frame-specific learnable latent code $\gamma$. The output of $\mathcal{T}_{\phi}$ is then processed by $\mathcal{D}_{\theta}$, a subsequent MLP network parameterized by $\theta$, which also takes the positional encoding of the viewing direction $\text{PE}(\vec{v})$ as input to predict the final RGB color $\mathbf{c}$ and volume density $\sigma$ for a given point in space. To compensate for inherent inaccuracies in facial expression and pose estimation during tracking, and to capture personalized dynamic details not fully encompassed by parametric models, our GAT-NeRF framework incorporates frame-specific learnable latent codes $\{\gamma_i\}_{i=1}^K$. Inspired by prior work~\cite{Gafni21DynamicNeural} demonstrating the efficacy of such latent codes in enhancing reconstruction fidelity, these codes are jointly optimized with all other GAT-NeRF parameters. They serve as a flexible, corrective mechanism, designed to dynamically rectify alignment discrepancies and to learn subtle facial variations arising from individual differences or complex expressions that are challenging for prior models like 3DMMs to represent accurately.

\begin{figure}[htb]
  \centering
  \includegraphics[width=\columnwidth]{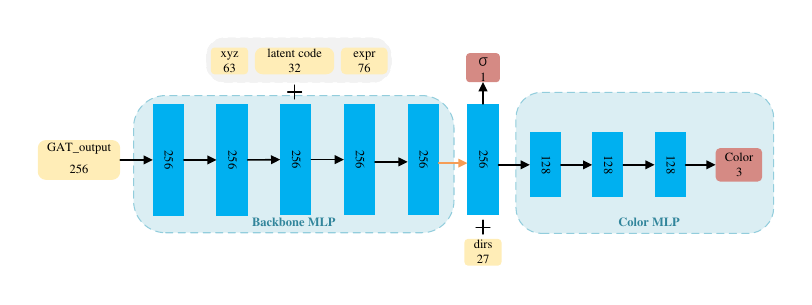}
  \Description{Architecture of the MLP networks following the GAT module in GAT-NeRF. The GAT output features are processed by a Backbone MLP, which includes a skip connection to fuse positional encoding, expression parameters, and latent codes with intermediate features. The backbone predicts volume density (sigma), while a separate Color MLP predicts RGB color (c) using view directions.}
  \caption{Architecture of the MLP networks following the GAT module in GAT-NeRF. The GAT output features are processed by a Backbone MLP, which incorporates a skip connection to fuse initial multi-modal inputs (positional encoding, expression, and latent codes) with intermediate features. This backbone subsequently predicts volume density ($\sigma$). The features are also used, along with view directions, by a separate Color MLP to predict RGB color ($\mathbf{c}$).}
  \label{fig:mlp_architecture}
\end{figure}

Our methodology establishes a synergistic bridge between the representational power of Transformer networks and the function approximation capabilities of MLPs to achieve high-fidelity facial reconstruction. The pipeline initiates by forming a rich, multi-modal feature vector through the concatenation of positionally-encoded spatial coordinates $\text{PE}(\mathbf{p})$, expression parameters $\delta$, and the latent code $\gamma$. This composite representation is subsequently transformed via a linear projection layer, mapping it into a 256-dimensional feature space conducive to processing by the GAT module. The architectural enhancements stem from two principal aspects: firstly, the GAT module facilitates a higher-order non-linear projection mechanism. This significantly augments the fusion of multi-modal features, enabling the effective integration of geometric (from $\text{PE}(\mathbf{p})$), expressive (from $\delta$), and latent (from $\gamma$) information streams. Secondly, the self-attention mechanism inherent to the Transformer architecture engenders an implicit dynamic feature re-weighting strategy. This allows the model to adaptively prioritize and refine local facial details, such as wrinkles and subtle skin marks, during the reconstruction process. This dual characteristic is pivotal for concurrently optimizing both the global facial structure and its intricate local details, a capability particularly advantageous for capturing the nuanced deformations characteristic of dynamic facial expressions. The attention mechanisms of the Transformer, therefore, operate in concert with the MLP's capacity for smooth interpolation, thereby maintaining spatial coherence while faithfully preserving individual-specific facial characteristics.

As illustrated in Figure 2, the 256-dimensional features enhanced by the GAT module are propagated into a backbone MLP network. This backbone network is comprised of five fully connected layers, each containing 256 neurons and employing ReLU activation functions. A significant architectural element is a skip connection introduced in the third layer of this backbone. At this juncture, the original concatenated input features (consisting of $\text{PE}(\mathbf{p})$, $\delta$, and $\gamma$), \textit{prior} to their processing by the GAT module, are concatenated with the GAT-processed features currently propagating through the backbone. This design choice serves to preserve raw input information and promote more effective feature fusion across different representational stages. The output from this backbone network is subsequently projected into a 256-dimensional feature vector, denoted as $\mathbf{feat}$. This vector, $\mathbf{feat}$, then serves as the input to a separate fully connected layer that predicts the volume density $\sigma$ for the sampled point. For color prediction, $\mathbf{feat}$ is concatenated with the positional encoding of the viewing direction $\text{PE}(\vec{v})$ (if view-dependent effects are enabled). This combined feature is then processed through a dedicated 4-layer MLP color prediction branch. The initial three layers of this branch consist of 128 neurons each with ReLU activation, while the final layer is a linear transformation that outputs the 3D RGB color values $\mathbf{c}$.

\subsection{Geometry-Aware Transformer (GAT) Module}
Our GAT module, whose core objective is to enhance the model's capacity for representing fine-grained geometric details, is not a novel Transformer architecture designed from scratch. Instead, it is built on the mature and standard transformer encoder layer~\cite{Vaswani17Attention} to ensure reliability and reproducibility. The central innovation of this method, therefore, lies not in modifying the internal mechanics of the Transformer, but in its unique application paradigm, which is specifically tailored for the task of reconstructing dynamic facial geometry. This paradigm integrates three key design considerations. First, for Unified Geometry-Aware Input Encoding, we implement a novel input fusion strategy that concatenates three critical information streams: (i) positionally-encoded 3D spatial coordinates $\text{PE}(\mathbf{p})$, providing high-frequency spatial context; (ii) facial expression parameters $\delta$ derived from a 3DMM, which explicitly encode geometric deformations of the face; and (iii) a learnable personalized latent code $\gamma$, which is designed to compensate for errors in expression parameter information and pose estimation. This unified input provides the model with comprehensive geometric priors, enabling it to capture both the global facial structure and subtle local deformations simultaneously. Furthermore, our approach features a Point-wise Feature Enhancement paradigm. In contrast to conventional applications of Transformers in sequence modeling or image patch processing, our GAT module employs this paradigm to process the features for each 3D sampled point independently (i.e., as a sequence of length one). This targeted approach allows the module to concentrate its representational power on refining the features of individual points, thereby significantly improving the reconstruction fidelity of local details, such as dynamic wrinkles, without incurring the computational cost of inter-point attention mechanisms. Finally, through Geometry-Informed Attention Learning, by incorporating explicit geometric priors (specifically, the expression parameters $\delta$ and the learned latent code $\gamma$) as integral components of the input features fed into the self-attention mechanism, the GAT module is guided to learn and emphasize feature dependencies relevant to dynamic facial geometry. This implicit guidance through geometry-rich inputs enhances the model's ability to reconstruct motion-dependent characteristics, such as expression-induced wrinkles and other subtle surface changes.

The hyperparameters of the GAT module are meticulously designed to form a lightweight yet effective component. We set the hidden dimension of the model ($d_{\text{model}}$) to 256 to align with the subsequent MLP layers for seamless feature integration, and the number of attention heads ($n_{\text{head}}$) to 8, a standard configuration for this dimension. Crucially, the decision to use a single transformer layer ($\textit{num\_layers}=1$) is a principled choice rooted in a rigorous cost-benefit analysis. Our primary design objective was to enhance the model's representational power for high-frequency details without introducing prohibitive computational overhead to the already intensive NeRF pipeline. As shown in Figure 3, a single Transformer encoder layer is sufficient to produce significant improvements in reconstruction quality, with a substantial reduction in L1 error and a notable increase in SSIM, which qualitatively translate to superior rendering of fine details like wrinkles and skin textures. This decision is further reinforced by its computational implications. Our analysis reveals that even with our targeted point-wise attention strategy, a single GAT layer results in a render time of approximately 24 seconds per frame (at a resolution of $512 \times 512$). While stacking more layers might offer marginal performance gains, it would lead to a super-linear increase in computational demand at the expense of practical usability. We therefore conclude that, for our application, a single-layer GAT represents the optimal trade-off point between reconstruction fidelity and computational efficiency.

Within our network architecture, the strategic integration of the GAT module with the subsequent MLP serves to ameliorate the reconstruction of high-frequency details. This is achieved through dynamic feature re-weighting facilitated by the self-attention mechanism and localized feature refinement via the non-linear transformations within the GAT. As notionally illustrated in Figure 1 , the GAT module executes its function through several critical processing stages: (a) concatenation of the multi-modal input features ($\text{PE}(\mathbf{p})$, $\delta$, $\gamma$); (b) dynamic feature re-weighting and contextualization via a multi-head self-attention mechanism; and (c) further non-linear feature enhancement using a position-wise feed-forward network.

The input features are initially concatenated as per Equation~\ref{eq:concat}:
\begin{equation}
X_{\text{concat}} = [X_{\text{coord}} \Vert X_{\text{expr}} \Vert X_{\text{latent}}] \in \mathbb{R}^{171}
\label{eq:concat}
\end{equation}
where $\text{PE}(\mathbf{p})$ utilizes $L_p=10$ frequency bands (resulting in a 63-dimensional vector), $\delta$ is 76-dimensional, and $\gamma$ is a 32-dimensional learnable latent code, yielding a total input dimension $D_{\text{in}} = 171$. These concatenated features $X_{\text{concat}}$ are then linearly projected into the Transformer's hidden space of dimension $D=256$: $X_{\text{proj}} = X_{\text{concat}}\mathbf{W}_p \in \mathbb{R}^{D}$, where $\mathbf{W}_p \in \mathbb{R}^{D_{\text{in}} \times D}$ is a learnable projection matrix.

Given the point-wise processing nature ($T=1$ effective sequence length), the input $X_{\text{proj}}$ to the Transformer layer can be viewed as having dimensions $(B, 1, D)$, where $B$ represents the batch size. The Query (Q), Key (K), and Value (V) matrices for the self-attention mechanism are computed from $X_{\text{proj}}$ as:

\begin{equation}
\begin{gathered}
Q = X_{\text{proj}} \mathbf{W}_Q, \quad K = X_{\text{proj}} \mathbf{W}_K, \quad V = X_{\text{proj}} \mathbf{W}_V, \\
\mathbf{W}_Q, \mathbf{W}_K, \mathbf{W}_V \in \mathbb{R}^{D \times d_k N_h}
\end{gathered}
\label{eq:qkv}
\end{equation}
where $N_h$ is the number of attention heads, and $d_k, d_v$ are the dimensions of keys/queries and values per head, respectively, such that $d_k N_h = d_v N_h = D$ is common. This point-wise operation allows the Transformer to focus on intra-feature interactions for each point, enhancing its representation for local detail modeling, while global scene consistency is enforced by the volumetric rendering process.

The multi-head self-attention mechanism computes $O_{\text{attn}} = \text{MultiHead}(Q, K, V)$. Following this, a position-wise Feed-Forward Network (FFN) processes the (layer-normalized) attention output to further refine features, particularly those pertinent to high-frequency details:

\begin{equation}
\begin{gathered}
\text{FFN}(O_{\text{norm1}}) = \text{ReLU}(O_{\text{norm1}}\mathbf{W}_1 + \mathbf{b}_1)\mathbf{W}_2 + \mathbf{b}_2, \\
\mathbf{W}_1 \in \mathbb{R}^{D \times D_{\text{ffn}}}, \mathbf{W}_2 \in \mathbb{R}^{D_{\text{ffn}} \times D}
\end{gathered}
\label{eq:ffn}
\end{equation}
where $D_{\text{ffn}}$ is the inner-layer dimensionality of the FFN (e.g., $D_{\text{ffn}}=2048$ for $D=256$).
The GAT module's final output, $\text{Output}_{\text{GAT}}$, is derived through a standard Transformer encoder layer architecture, incorporating two residual connections and layer normalizations:
\begin{equation}
\begin{aligned}
O'_{\text{attn}} &= X_{\text{proj}} + \text{MultiHeadAttention}(Q, K, V) \\
O_{\text{norm1}} &= \text{LayerNorm}(O'_{\text{attn}}) \\
O'_{\text{ffn}} &= O_{\text{norm1}} + \text{FFN}(O_{\text{norm1}}) \\
\text{Output}_{\text{GAT}} &= \text{LayerNorm}(O'_{\text{ffn}})
\end{aligned}
\end{equation}

These architectural components (residual connections and layer normalizations) are instrumental in stabilizing the training dynamics of deep networks, balancing the propagation of detailed information with global contextual cues, and effectively adapting to the point-wise processing regime. They ensure that the original input information ($X_{\text{proj}}$) is adequately preserved and integrated during the deep feature transformations aimed at modeling local details.

\subsection{Density and Color Prediction via Volumetric Rendering}
The synthesis of 2D images from our implicit neural representation is achieved through a differentiable volume rendering pipeline. For any given 3D point $\mathbf{r}(t) = \mathbf{o} + t\mathbf{d}$ sampled along a camera ray (where $\mathbf{o}$ is the camera origin, $\mathbf{d}$ is the unit viewing direction vector, and $t$ is the depth parameter varying between near $t_{\text{near}}$ and far $t_{\text{far}}$ clipping planes), our GAT-NeRF network (defined by $\mathcal{T}_{\phi}$ and $\mathcal{D}_{\theta}$ as per Eq.~\ref{eq:model}) predicts its volume density $\sigma(\mathbf{r}(t))$ and view-dependent color $\mathbf{c}(\mathbf{r}(t), \mathbf{d})$. The final color $\hat{C}(\mathbf{r})$ for the pixel corresponding to ray $\mathbf{r}$ is then computed by numerically integrating these quantities along the ray:
\begin{equation}
    \hat{C}(\mathbf{r}) = \int_{t_{\text{near}}}^{t_{\text{far}}} T(t) \sigma(\mathbf{r}(t)) \mathbf{c}(\mathbf{r}(t), \mathbf{d}) \, dt
    \label{eq:volume_integration_final} 
\end{equation}
where $T(t)$ represents the accumulated transmittance from $t_{\text{near}}$ to $t$, indicating the probability that the ray travels from $t_{\text{near}}$ to $t$ without being occluded:
\begin{equation}
    T(t) = \exp\left( -\int_{t_{\text{near}}}^{t} \sigma(\mathbf{r}(s)) \, ds \right)
    \label{eq:transmittance_final} 
\end{equation}

In practice, the continuous integrals in Eq.~\ref{eq:volume_integration_final} and Eq.~\ref{eq:transmittance_final} are approximated using quadrature rules. We employ the hierarchical volumetric sampling strategy introduced by Mildenhall et al.~\cite{Mildenhall21Nerf}. This involves training two instances of our GAT-NeRF network in parallel: a "coarse" network, parameterized by $\theta_c$, and a "fine" network, parameterized by $\theta_f$. While both networks share the identical GAT-NeRF architecture, they possess independent sets of learnable weights. During rendering, $N_c$ points are initially sampled uniformly along each ray. The density predictions from the coarse network $\sigma_c(\mathbf{r}(t))$ are utilized to form a piecewise-constant probability distribution along the ray. Subsequently, an additional $N_f$ points are sampled based on this distribution, effectively concentrating samples in regions of higher expected scene content. The final rendered color for the ray is then obtained by processing all $N_c + N_f$ sampled points through the fine network.

\subsection{Training Objectives and Implementation Details}
The optimization of our GAT-NeRF model, encompassing the parameters of both the coarse ($\theta_c$) and fine ($\theta_f$) networks, as well as the set of learnable latent codes $\{\gamma_i\}$, is driven by a photometric reconstruction loss. Additionally, to prevent overfitting and encourage smoother latent representations, we regularize the frame-specific latent codes $\gamma_i \in \mathbb{R}^{32}$ using an $\ell_2$ penalty. The total loss function $\mathcal{L}_{\text{total}}$ is formulated as:

\begin{align}
  \mathcal{L}_{\text{total}} &= \sum_{i=1}^{K} \left[
    \mathcal{L}_{\text{photo},i}(\theta_c, \gamma_i)
    + \mathcal{L}_{\text{photo},i}(\theta_f, \gamma_i)
    + \lambda_{\gamma} \|\gamma_i\|^2_2
  \right]
  \label{eq:total_loss_final_with_reg} \\[1ex]
  \mathcal{L}_{\text{photo},i}(\theta, \gamma_i) &= 
  \sum_{\mathbf{r} \in \mathcal{R}_i} \left\|
    \hat{C}(\mathbf{r}; \theta, P_i, \delta_i, \gamma_i)
    - C^{\text{gt}}(\mathbf{r})
  \right\|^2_2
  \label{eq:photometric_loss_final}
\end{align}
where $K$ denotes the total number of frames in the training set, $\mathcal{R}_i$ is the batch of rays sampled for the $i$-th training frame, and $P_i, \delta_i, \gamma_i$ are the respective rigid head pose, expression parameters, and learnable latent code for frame $i$. $C^{\text{gt}}(\mathbf{r})$ represents the ground truth RGB color of the pixel corresponding to the ray $\mathbf{r}$. The term $\mathcal{L}_{\text{photo},i}$ denotes the photometric reconstruction loss for frame $i$ using network parameters $\theta$ (either $\theta_c$ or $\theta_f$) and the specific latent code $\gamma_i$. The hyperparameter $\lambda_{\gamma}$ controls the strength of the $\ell_2$ regularization on the latent codes, which we set to $0.05$ in our experiments. The notation $\mathcal{L}_{\text{photo},i}(\theta, \gamma_i)$ explicitly indicates that the photometric loss for a given frame is a function of both the network weights $\theta$ and the per-frame latent code $\gamma_i$, as $\gamma_i$ is an input to the model (Eq.~\ref{eq:model}) and thus influences the rendered color $\hat{C}$. Both $\theta$ and all $\{\gamma_i\}$ are optimized jointly.

During each training iteration, for a given input frame $I_i$, a batch of 1024 rays is randomly sampled. To focus computational resources on the primary subject, most rays are sampled from within facial bounding boxes, which are derived from the parametric deformation model. The per-ray sampling strategy adheres to the two-stage hierarchical approach: $N_c=64$ points are initially sampled uniformly along each ray for the coarse network $D_c$. Based on the density distribution predicted by $D_c$, an additional $N_f=64$ points are importance-sampled for the fine network $D_f$. The final color prediction for each ray is then derived from these $N_c+N_f$ points using $D_f$.

The GAT-NeRF framework is implemented using the PyTorch library~\cite{Paszke19Pytorch}. The parameters of the coarse and fine networks ($\theta_c, \theta_f$), along with the set of learnable latent codes $\{\gamma_i\}$, are jointly optimized using the Adam optimizer~\cite{Kingma14Adam}. We employ a learning rate of $3 \times 10^{-4}$. Each model configuration is trained for 300,000 iterations, utilizing input images at a resolution of $512 \times 512$ pixels.

\begin{figure}[htb]
  \centering
  \includegraphics[width=1\columnwidth]{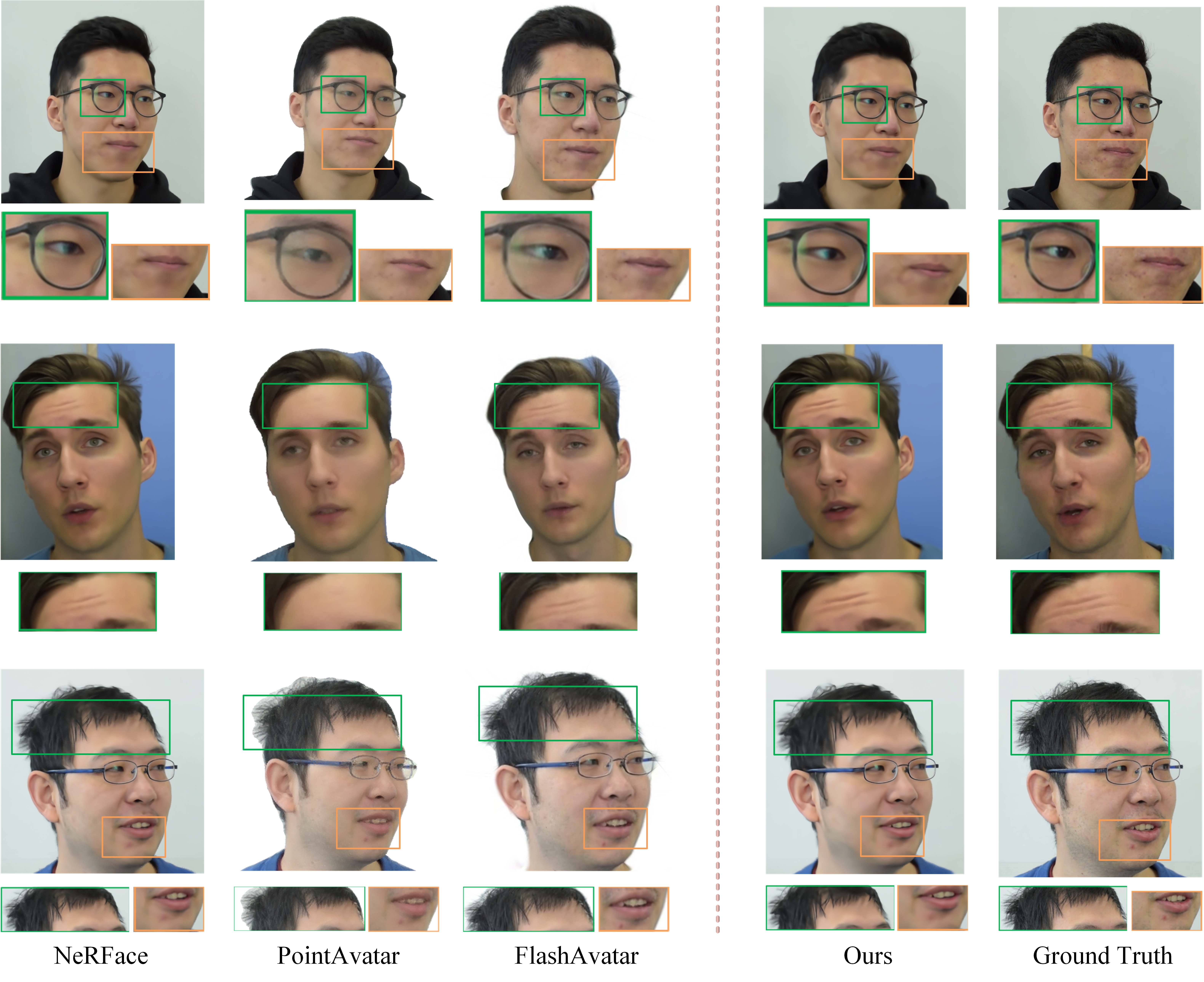}
  \Description{Qualitative comparison between GAT-NeRF and prior face reconstruction methods: NeRFace, PointAvatar, FlashAvatar, and ground truth. GAT-NeRF achieves better reconstruction of fine facial details such as wrinkles around the eyes and mouth.}
  \caption{We qualitatively compare our GAT-NeRF approach with classical and state-of-the-art face reconstruction methods. From left to right: NeRFace~\cite{Gafni21DynamicNeural}, PointAvatar~\cite{Zheng23Pointavatar}, FlashAvatar~\cite{Xiang24Flashavatar}, the proposed GAT-NeRF, and the ground-truth image. Zoomed-in regions highlight differences in fine detail preservation, particularly around the eyes, mouth, and forehead, where our method demonstrates superior reconstruction of dynamic wrinkles and subtle textures.}
  \label{fig:qualitative_comparison}
\end{figure}

\section{Experimental Evaluation}
This section details our experimental setup and presents comprehensive qualitative and quantitative evaluations of GAT-NeRF in terms of facial reconstruction, as well as expression and pose control. In addition, we conducted ablation studies to verify the necessity and contribution of each key component in our proposed GAT-NeRF architecture.

\subsection{Experimental Setup}
To make a direct and fair comparison with previous studies, we use the publicly available NeRFace dataset~\cite{Gafni21DynamicNeural} as it is a standard benchmark to evaluate such specific topic models. This dataset comprises monocular RGB video sequences (approximately 1-2 minutes per subject) capturing a diverse range of rigid and non-rigid facial movements. All video frames were cropped and resized to a resolution of $512 \times 512$ pixels to ensure comprehensive coverage of the facial region. For each subject's data, we performed a chronological split, with the first 90\% of the frames designated as the training set and the last 10\% of the frames reserved as the test set. During the recording sessions, subjects engaged in natural conversation, exhibiting a variety of facial expressions, including smiles, mouth openings, and other articulation-related movements, thereby providing rich data for evaluating dynamic facial reconstruction performance. Our model training and the evaluation of all baseline methods were performed on a consistent hardware platform to ensure fair comparisons.

\begin{table}[htbp]
\centering
\caption{Quantitative comparison with state-of-the-art methods.}
\label{tab:comparison}
\begin{tabular}{lcccc}
\toprule
\textbf{Method} 
 & \textbf{L1} $\downarrow$
 & \textbf{PSNR} $\uparrow$
 & \textbf{SSIM} $\uparrow$
 & \textbf{LPIPS} $\downarrow$ \\
\midrule
NeRFace~\cite{Gafni21DynamicNeural}   & 0.047 & 24.092 & 0.926 & 0.074 \\
PointAvatar~\cite{Zheng23Pointavatar} & 0.015 & \textbf{27.000} & 0.915 & \textbf{0.069} \\
FlashAvatar~\cite{Xiang24Flashavatar} & 0.015 & 26.883 & 0.916 & 0.071 \\
Ours                                  & \textbf{0.013} & 24.822 & \textbf{0.932} & 0.070 \\
\bottomrule
\end{tabular}
\end{table}

\subsection{Reconstruction and Expression-Driven Results}
We compare GAT-NeRF against several classical and state-of-the-art facial reconstruction methods. These include: NeRFace~\cite{Gafni21DynamicNeural}, PointAvatar~\cite{Zheng23Pointavatar}, and the 3DGS-based FlashAvatar~\cite{Xiang24Flashavatar}. Figure 3 presents qualitative comparisons in self-reenactment scenarios. As illustrated, our GAT-NeRF method preserves superior facial details compared to existing approaches. Notably, it excels in accurately reproducing realistic forehead wrinkles and perioral acne scars, details that are often over-smoothed or entirely missed by other methods. This enhanced capability is attributed to the effective processing and feature enhancement of multi-modal inputs including geometric priors by the GAT-NeRF architecture.

As shown in Table~\ref{tab:comparison}, we evaluate our method against the state-of-the-art approaches using four established metrics: L1 Loss($\downarrow$), Peak Signal-to-Noise Ratio (PSNR$\uparrow$), Structural Similarity Index (SSIM$\uparrow$)~\cite{Wang04ImageQuality}, and the perceptual LPIPS metric($\downarrow$)~\cite{Zhang18UnreasonableEffectiveness}. The quantitative results highlight the compelling performance of GAT-NeRF. Our method achieves the best scores for L1 loss and SSIM, demonstrating its superior capability to reconstruct fine-grained structures and maintaining high structural fidelity. This substantiates the efficacy of our GAT module in enhancing the representation of local geometric patterns, leading to precise recovery of details like wrinkles and skin textures. Furthermore, GAT-NeRF achieves a highly competitive LPIPS score of 0.070, nearly matching the leading method and indicating that our reconstructions are perceptually very close to the ground truth.

It should be noted that while our model does not attain the highest PSNR, its leading performance in structure-focused metrics (L1 and SSIM) suggests a deliberate trade-off. Our framework prioritizes the faithful reconstruction of visually significant details over optimizing for pixel-level average fidelity, to which PSNR is particularly sensitive. This design choice, aimed at achieving higher perceptual realism, is further validated and explained in our ablation study (Section~\ref{sec:ablation_studies}).

\begin{figure}[htb]
  \centering
  \includegraphics[width=1\columnwidth]{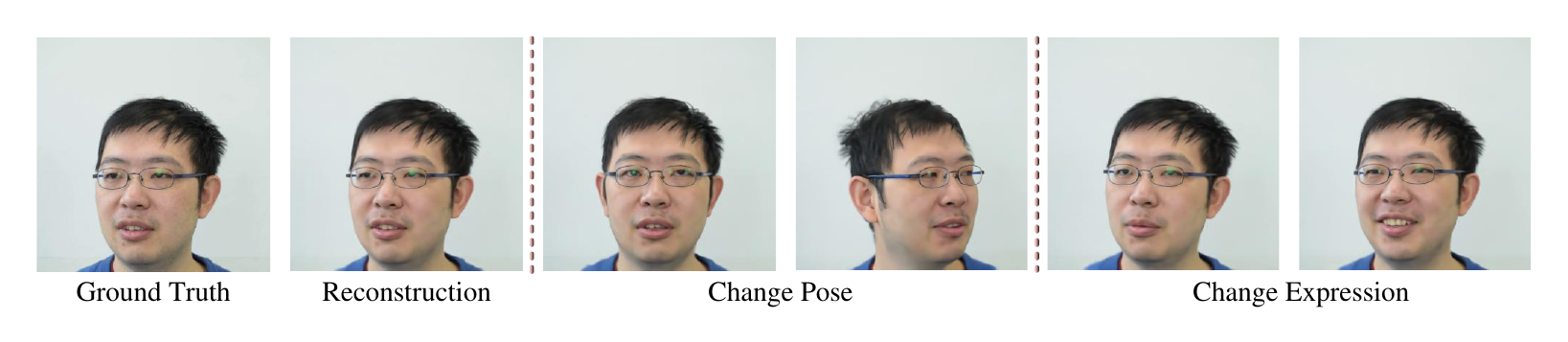}
  \Description{The dynamic neural radiance field enables independent manipulation of facial expression vectors and head poses.}
  \caption{Our dynamic neural radiance field enables decoupled control of expression vectors and head poses.}
  \label{fig:decoupled_control}
\end{figure}

\begin{figure}[htb]
  \centering
  \includegraphics[width=0.7\columnwidth]{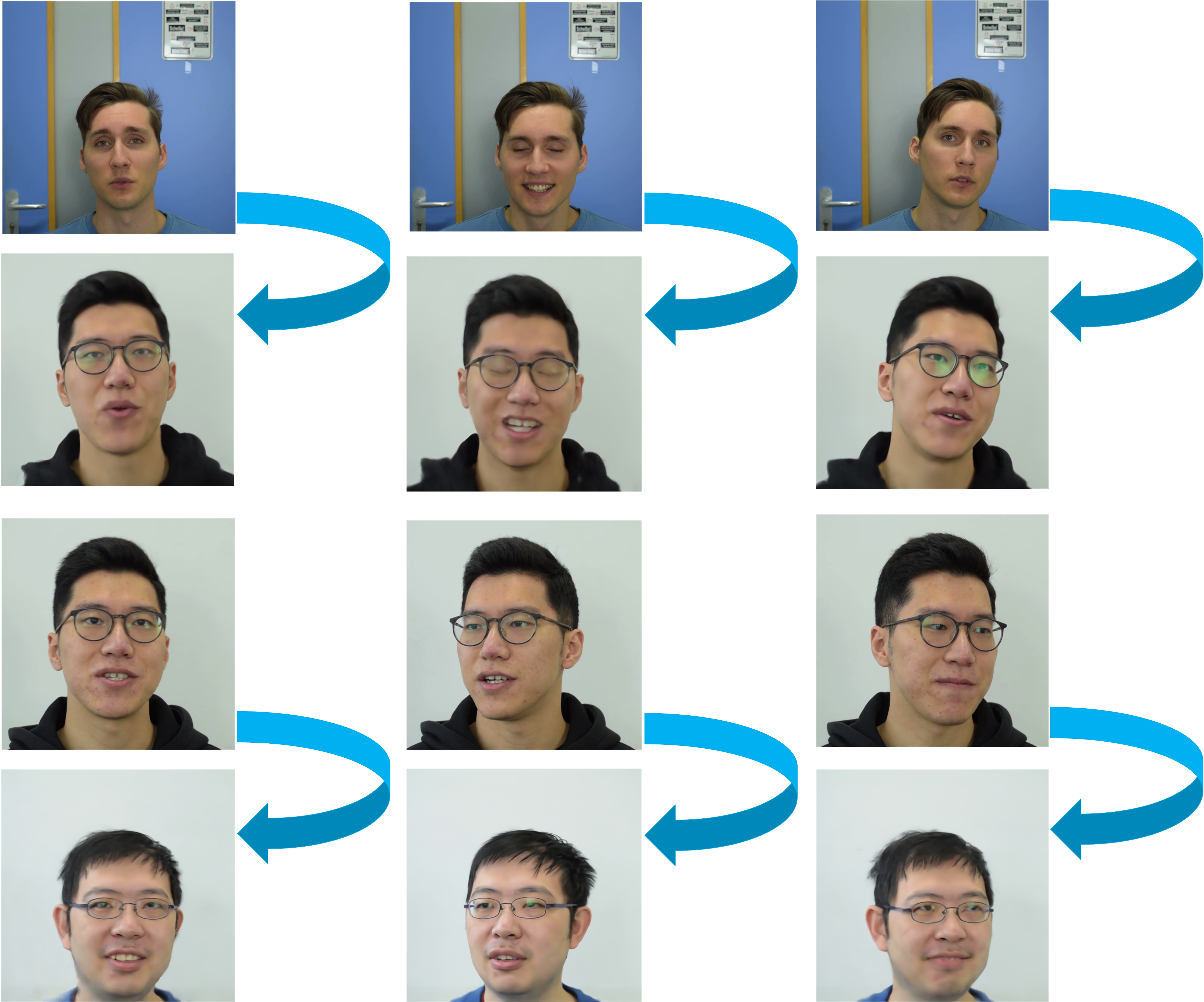}
  \Description{Demonstration of 4D facial avatar enabling facial expression reenactment. The expressions of a source subject are transferred to a target subject represented by a dynamic neural radiance field.}
  \caption{Our 4D facial avatar enables facial expression reenactment, where the expressions of a source subject are transferred to a target subject represented by our dynamic neural radiance field.}
  \label{fig:reenactment}
\end{figure}

\subsection{Expression and Pose Control}
GAT-NeRF models the dynamic radiance field by learning disentangled expression parameters $\delta$ and latent codes $\gamma_i$ during training. This enables the reconstruction of high-fidelity 4D facial models and, subsequently, precise control over expressions and poses. 

Figure 4 illustrates the capability of our model, GAT-NeRF, for decoupled control over expression and pose. From left to right: The ground truth image and its high-fidelity reconstruction. Next, we show novel pose synthesis by manipulating the head pose parameters while keeping the expression fixed. Finally, we present novel expression synthesis by altering the expression parameters while maintaining a fixed head pose. This illustrates GAT-NeRF's capability to independently control facial expressions and rigid head motion. We also present results of facial expression transfer results (Figure 5), which transfer expressions between different individuals. These experimental results indicate that our dynamic radiance field effectively captures and represents the speaker's appearance and geometry, allowing for flexible semantic editing.

\begin{figure}[htb]
  \centering
  \includegraphics[width=1\columnwidth]{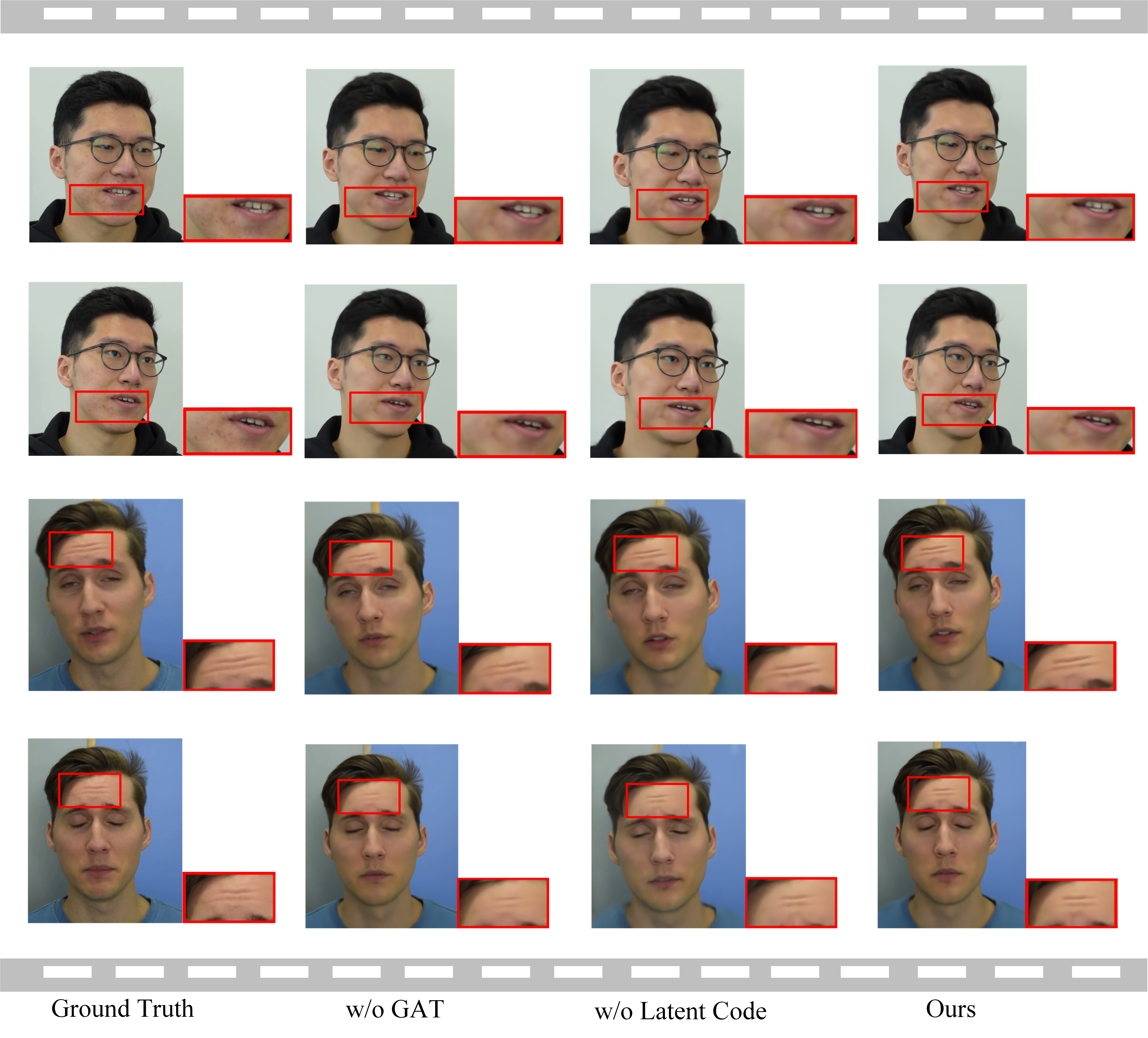}

  \Description{Ablation study on the key components of our model. Zoom-in views on the mouth and forehead regions show how both the GAT module and the latent code contribute to a better reconstruction of high-frequency facial details.}
  \caption{Ablation study on the key components of our model. Zoom-in views on the mouth and forehead regions show how both the GAT module and the latent code contribute to a better reconstruction of high-frequency facial details.}
  \label{fig:gat_ablation}
\end{figure}

\subsection{Ablation Studies}
\label{sec:ablation_studies}
To systematically evaluate the necessity and contribution of each key component in our proposed GAT-NeRF architecture, we conducted a comprehensive ablation study. This study compares three core model configurations, as detailed in Table~\ref{tab:ablation}: 
a) Baseline NeRF (w/o GAT), which uses a conventional MLP architecture without our GAT module; 
b) GAT-NeRF (w/o Latent Code), which incorporates the GAT module but removes the personalized latent code; 
and c) GAT-NeRF (Full Model), our final proposed architecture. By comparing these configurations, we can deconstruct and quantify the respective contributions of the GAT module and the latent code.

\begin{table}[!t]
\caption{Quantitative ablation analysis on the contributions of the Latent Code and GAT modules.}
\label{tab:ablation}
\centering
\begin{tabular}{lcccc}
\toprule
Method 
 & \textbf{L1} $\downarrow$
 & \textbf{PSNR} $\uparrow$
 & \textbf{SSIM} $\uparrow$
 & \textbf{LPIPS} $\downarrow$ \\
\midrule
w/o GAT & 0.047 & 24.092 & 0.926 & 0.074 \\
w/o Latent Code & \textbf{0.012} & \textbf{25.784} & \textbf{0.936} & 0.080 \\
Ours     & 0.013 & 24.822 & 0.932 & \textbf{0.070} \\
\bottomrule
\end{tabular}
\end{table}

Figure 6 presents a qualitative comparison of the reconstruction results. It is clearly observable that the full GAT-NeRF model demonstrates superior ability in reconstructing high-frequency facial details, such as dynamic wrinkles and acne marks. The quantitative results in Table~\ref{tab:ablation} not only corroborate these visual observations but also reveal insightful details about each component's role.

The quantitative results reported in Table~\ref{tab:ablation} provide strong empirical support for our qualitative observations and reveal insightful details about each component's contribution. We conducted a step-wise ablation analysis to dissect these effects. First, to isolate the efficacy of the GAT module, we compared the baseline NeRF (the w/o GAT setting) with the variant that only includes our GAT module (w/o Latent Code). The introduction of the GAT module yields a dramatic improvement across all metrics: L1 error plummets from 0.047 to 0.012, PSNR increases significantly from 24.092 to 25.784~dB, and SSIM improves to 0.936. This result unequivocally demonstrates that the GAT module, by performing point-wise feature enhancement on multi-modal geometric inputs, is the cornerstone of our framework's high-fidelity reconstruction capabilities. Next, we analyzed the role of the personalized latent code by comparing the w/o Latent Code variant with our full model. This comparison uncovers a classic trade-off between pixel-level accuracy and perceptual realism. While the model without the latent code achieves slightly better scores on pixel-based metrics (L1, PSNR, and SSIM), our full model obtains a significantly superior LPIPS score (0.070 vs. 0.080). This indicates that the latent code, while introducing minor pixel-level deviations, guides the network to synthesize finer, more realistic high-frequency details, such as dynamic wrinkles and skin textures. These details are better aligned with human perception, a quality effectively captured by the LPIPS metric, but sometimes penalized by PSNR, which favors overly smooth results. This quantitative finding is strongly corroborated by the qualitative results in Figure 6, where the full model visibly reconstructs more intricate facial details.

In summary, this step-wise analysis demonstrates that the GAT module is essential for establishing a high-quality baseline reconstruction, while the latent code acts as a crucial refinement vector. It compensates for imperfections in the 3DMM priors and pushes the model beyond simple pixel-wise regression towards generating truly high-fidelity, perceptually convincing facial avatars. The synergy between these two components drives the cumulative performance of our proposed method.

\section{Discussion}
Although the proposed GAT-NeRF method is capable of producing high-fidelity head reconstructions from monocular videos and demonstrates superior performance in capturing fine dynamic details, several limitations and avenues for future research remain.

Firstly, a primary limitation of our current approach lies in its generalization capability, a challenge it shares with many state-of-the-art neural avatar systems. GAT-NeRF operates under a per-subject optimization paradigm. Although this enables exceptional detail fidelity for a specific individual, it inherently restricts its direct applicability to unseen subjects. This dependency on subject-specific training poses a significant challenge for scenarios that require rapid adaptation to new users and highlights a critical open research problem in the field. 
However, we believe that our work offers a valuable component in addressing this challenge. By demonstrating a robust method for capturing high-frequency, geometry-aware details, the GAT module can serve as a powerful detail-enhancement cornerstone for future, more generalizable architectures. A particularly promising direction involves integrating our approach with advanced 3D-aware generative models. Future work could explore learning a generalizable facial prior, upon which a GAT-like mechanism could then be leveraged to inject personalized dynamic details. Such a hybrid strategy has the potential to bridge the gap between the high fidelity of subject-specific models and the flexibility of generalizable ones, paving the way for the creation of scalable and practical digital humans while reducing the reliance on extensive per-subject data.

Secondly, while our point-wise GAT module design enhances feature representation while maintaining computational efficiency relative to more complex Transformer architectures, NeRF-based solutions generally still exhibit a gap in training and rendering speeds when compared to the latest explicit representation methods, such as 3DGS. Exploring hybrid representations that combine the feature enhancement strengths of our GAT module with more efficient rendering pipelines like those used in 3DGS, or developing novel hybrid GAT-3DGS architectures, could be a promising direction to achieve both high-fidelity detail and real-time performance.

Furthermore, the latent codes \(\gamma_i\), while effective in compensating for tracking inaccuracies and capturing personalized details, offer limited interpretability. Investigating methods to disentangle these latent codes into more semantically meaningful control parameters, such as specific micro-expressions or finer-grained identity characteristics, would significantly enhance the model's controllability and practical utility.

\section{Conclusion}
In this paper, we have presented GAT-NeRF, a novel hybrid neural radiance field framework for high-fidelity, controllable 4D facial avatar reconstruction from monocular video streams. Our framework innovatively combines a coordinate-aligned MLP with a lightweight, GAT module. By effectively leveraging prior knowledge derived from a 3DMM in the form of expression parameters $\delta$, along with frame-specific learnable latent codes $\gamma_i$, as geometry-aware inputs to the GAT module, our method significantly improves the capability to capture and reconstruct fine-grained facial details, such as dynamic wrinkles and personalized textures.

Distinct from traditional NeRF approaches that rely solely on MLPs to regress the volumetric radiance field, GAT-NeRF overcomes the limitations of MLPs in modeling high-frequency variations by introducing a point-wise GAT module. This module performs local feature enhancement and dynamic re-weighting on multi-modal features that fuse spatial, expression, and latent code information, achieved through its internal self-attention mechanism and feed-forward networks. This design enhances feature expressiveness via higher-order non-linear projections and attention-driven refinement without incurring the high computational cost of inter-point interactions, thereby maintaining relative computational efficiency. Our work establishes a new direction for neural facial avatar creation, achieving an effective balance between geometric precision and photorealistic detail. Future research endeavors will focus on enhancing the generalization capabilities of neural radiance field representations and reducing their dependence on subject-specific training data to facilitate their deployment in a broader range of practical applications.

\begin{acks}
This work was funded by UKRI (EP/Z000025/1) and the Horizon Europe Programme under the MSCA grant for ACMod (Grant No. 101130271).
\end{acks}

\bibliographystyle{ACM-Reference-Format}
\bibliography{sample-base} 

@article{xia2022aflfp,
  title   = {{AFLFP}: A database with annotated facial landmarks for facial palsy},
  author  = {Xia, Yifan and Nduka, Charles and Kannan, Ruben Yap and Pescarini, Elena and Berner, Juan Enrique and Yu, Hui},
  journal = {IEEE Transactions on Computational Social Systems},
  volume  = {10},
  number  = {4},
  pages   = {1975--1985},
  year    = {2022}
}

@article{cao2023high,
  title   = {Is high-fidelity important for human-like virtual avatars in human computer interactions?},
  author  = {Cao, Qiongdan and Yu, Hui and Charisse, Paul and Qiao, Si and Stevens, Brett},
  journal = {International Journal of Network Dynamics and Intelligence},
  volume  = {2},
  number  = {1},
  pages   = {15--23},
  year    = {2023}
}

@article{wang2023mgeed,
  title   = {{Mgeed}: A multimodal genuine emotion and expression detection database},
  author  = {Wang, Yiming and Yu, Hui and Gao, Weihong and Xia, Yifan and Nduka, Charles},
  journal = {IEEE Transactions on Affective Computing},
  volume  = {15},
  number  = {2},
  pages   = {606--619},
  year    = {2023}
}

@inproceedings{Ge25Implicit,
  author    = {Ge, Shiping and Chen, Qiang and Jiang, Zhiwei and Yin, Yafeng and Qin, Liu and Chen, Ziyao and Gu, Qing},
  title     = {Implicit Location-Caption Alignment via Complementary Masking for Weakly-Supervised Dense Video Captioning},
  booktitle = {Proceedings of the {AAAI} Conference on Artificial Intelligence},
  volume    = {39},
  number    = {3},
  year      = {2025},
  pages     = {3113--3121}
}

@article{Qiao25Multimodal,
  author    = {Qiao, Yu and Lu, Wei and Jing, Peiguang and Wang, Weiming and Su, Yuting},
  title     = {Multimodal Dual-Graph Collaborative Network With Serial Attentive Aggregation Mechanism for Micro-Video Multi-Label Classification},
  journal   = {{IEEE} Transactions on Multimedia},
  year      = {2025},
  note      = {To appear}
}

@article{Liu24Enhancing,
  author    = {Liu, Weijia and Cao, Jiuxin and Wei, Ran and Zhu, Xuelin and Liu, Bo},
  title     = {Enhancing micro-video venue recognition via multi-modal and multi-granularity object relations},
  journal   = {{IEEE} Transactions on Circuits and Systems for Video Technology},
  volume    = {34},
  number    = {7},
  year      = {2024},
  pages     = {5440--5451}
}

@article{Jing24Multimodal,
  author    = {Jing, Peiguang and Liu, Xianyi and Zhang, Lijuan and Li, Yun and Liu, Yu and Su, Yuting},
  title     = {Multimodal attentive representation learning for micro-video multi-label classification},
  journal   = {{ACM} Transactions on Multimedia Computing, Communications and Applications},
  volume    = {20},
  number    = {6},
  year      = {2024},
  pages     = {1--23}
}

@article{Thai25SplatTalk,
  author    = {Thai, Anh and Peng, Songyou and Genova, Kyle and Guibas, Leonidas and Funkhouser, Thomas},
  title     = {{SplatTalk: 3D VQA with Gaussian Splatting}},
  journal   = {arXiv preprint arXiv:2503.06271},
  year      = {2025}
}

@article{Peng243DVLGS,
  author    = {Peng, Qucheng and Planche, Benjamin and Gao, Zhongpai and Zheng, Meng and Choudhuri, Anwesa and Chen, Terrence and Chen, Chen and Wu, Ziyan},
  title     = {{3D} Vision-Language Gaussian Splatting},
  journal   = {arXiv preprint arXiv:2410.07577},
  year      = {2024}
}

@article{Jiang25AnySplat,
  author    = {Jiang, Lihan and Mao, Yucheng and Xu, Linning and Lu, Tao and Ren, Kerui and Jin, Yichen and Xu, Xudong and et al.},
  title     = {{AnySplat: Feed-forward 3D Gaussian Splatting from Unconstrained Views}},
  journal   = {arXiv preprint arXiv:2505.23716},
  year      = {2025}
}

@inproceedings{Ma21Pixel,
  author    = {Ma, Shugao and Simon, Tomas and Saragih, Jason and Wang, Dawei and Li, Yuecheng and De La Torre, Fernando and Sheikh, Yaser},
  title     = {{Pixel codec avatars}},
  booktitle = {Proceedings of the {IEEE/CVF} Conference on Computer Vision and Pattern Recognition},
  year      = {2021},
  pages     = {64--73}
}

@inproceedings{Wang21Ibrnet,
  author    = {Wang, Qianqian and Wang, Zhicheng and Genova, Kyle and Srinivasan, Pratul P. and Zhou, Howard and Barron, Jonathan T. and Martin-Brualla, Ricardo and Snavely, Noah and Funkhouser, Thomas},
  title     = {{Ibrnet}: Learning multi-view image-based rendering},
  booktitle = {Proceedings of the {IEEE/CVF} Conference on Computer Vision and Pattern Recognition},
  year      = {2021},
  pages     = {4690--4699}
}

@inproceedings{Johari22Geonerf,
  author    = {Johari, Mohammad Mahdi and Lepoittevin, Yann and Fleuret, Fran{\c{c}}ois},
  title     = {{Geonerf}: Generalizing {NeRF} with geometry priors},
  booktitle = {Proceedings of the {IEEE/CVF} Conference on Computer Vision and Pattern Recognition},
  year      = {2022},
  pages     = {18365--18375}
}

@inproceedings{Chen21Mvsnerf,
  author    = {Chen, Anpei and Xu, Zexiang and Zhao, Fuqiang and Zhang, Xiaoshuai and Xiang, Fanbo and Yu, Jingyi and Su, Hao},
  title     = {{Mvsnerf}: Fast generalizable radiance field reconstruction from multi-view stereo},
  booktitle = {Proceedings of the {IEEE/CVF} International Conference on Computer Vision},
  year      = {2021},
  pages     = {14124--14133}
}

@article{Sun24MultiModalDriven,
  author    = {Sun, Kuiyuan and Liu, Xiaolong and Li, Xiaolong and Zhao, Yao and Wang, Wei},
  title     = {Multi-Modal Driven Pose-Controllable Talking Head Generation},
  journal   = {{ACM} Transactions on Multimedia Computing, Communications and Applications},
  volume    = {20},
  number    = {12},
  year      = {2024},
  articleno = {1},
  numpages  = {23}
}

@article{Zou24_4DFacialExpression,
  author    = {Zou, Kaifeng and Faisan, Sylvain and Yu, Boyang and Valette, S{\'{e}}bastien and Seo, Hyewon},
  title     = {{4D} facial expression diffusion model},
  journal   = {{ACM} Transactions on Multimedia Computing, Communications and Applications},
  volume    = {21},
  number    = {1},
  year      = {2024},
  articleno = {1},
  numpages  = {23}
}

@article{Liu24MultimodalFusion,
  author    = {Liu, Zhilei and Liu, Xiaoxing and Chen, Sen and Liu, Jiaxing and Wang, Longbiao and Bi, Chongke},
  title     = {Multimodal Fusion for Talking Face Generation Utilizing Speech-Related Facial Action Units},
  journal   = {{ACM} Transactions on Multimedia Computing, Communications and Applications},
  volume    = {20},
  number    = {9},
  year      = {2024},
  articleno = {1},
  numpages  = {24}
}

@article{Liu23TalkingFaceAnatomy,
  author    = {Liu, Shiguang and Wang, Huixin},
  title     = {Talking face generation via facial anatomy},
  journal   = {{ACM} Transactions on Multimedia Computing, Communications and Applications},
  volume    = {19},
  number    = {3},
  year      = {2023},
  articleno = {1},
  numpages  = {19}
}

@article{Shamshiri24TextMining,
  author    = {Shamshiri, Alireza and Ryu, Kyeong Rok and Park, June Young},
  title     = {Text mining and natural language processing in construction},
  journal   = {Automation in Construction},
  volume    = {158},
  year      = {2024},
  pages     = {105200}
}

@article{Zhao24AutonomousDriving,
  author    = {Zhao, Jingyuan and Zhao, Wenyi and Deng, Bo and Wang, Zhenghong and Zhang, Feng and Zheng, Wenxiang and Cao, Wanke and Nan, Jinrui and Lian, Yubo and Burke, Andrew F.},
  title     = {Autonomous driving system: A comprehensive survey},
  journal   = {Expert Systems with Applications},
  volume    = {242},
  year      = {2024},
  pages     = {122836}
}

@article{Zhou24AdaptiveSegmentation,
  author    = {Zhou, Xiaokang and Liang, Wei and Kawai, Akira and Fueda, Kaoru and She, Jinhua and Wang, Kevin I-Kai},
  title     = {Adaptive segmentation enhanced asynchronous federated learning for sustainable intelligent transportation systems},
  journal   = {{IEEE} Transactions on Intelligent Transportation Systems},
  volume    = {25},
  number    = {7},
  year      = {2024},
  pages     = {6658--6666}
}

@article{Bakirci24EnhancingVehicle,
  author    = {Bakirci, Murat},
  title     = {Enhancing vehicle detection in intelligent transportation systems via autonomous {UAV} platform and {YOLOv8} integration},
  journal   = {Applied Soft Computing},
  volume    = {164},
  year      = {2024},
  pages     = {112015}
}

@inproceedings{Mazzacca23NeRFHeritage,
  author    = {Mazzacca, G. and Karami, A. and Rigon, S. and Farella, E. M. and Trybala, P. and Remondino, F.},
  title     = {{NeRF} for heritage {3D} reconstruction},
  booktitle = {International Archives of the Photogrammetry, Remote Sensing and Spatial Information Sciences},
  volume    = {XLVIII-M-2-2023},
  year      = {2023},
  pages     = {1051--1058}
}

@article{Zhang20NeRFplusplus,
  author    = {Zhang, Kai and Riegler, Gernot and Snavely, Noah and Koltun, Vladlen},
  title     = {{NeRF++}: Analyzing and improving neural radiance fields},
  year      = {2020},
  journal   = {arXiv preprint arXiv:2010.07492}
}

@inproceedings{Lin20SDFSRN,
  author    = {Lin, Chen-Hsuan and Wang, Chaoyang and Lucey, Simon},
  title     = {{SDF-SRN}: Learning signed distance {3D} object reconstruction from static images},
  booktitle = {Advances in Neural Information Processing Systems},
  volume    = {33},
  year      = {2020},
  pages     = {11453--11464}
}

@article{Kato20DifferentiableRenderingSurvey,
  author    = {Kato, Hiroharu and Beker, Deniz and Morariu, Mihai and Ando, Takahiro and Matsuoka, Toru and Kehl, Wadim and Gaidon, Adrien},
  title     = {Differentiable Rendering: A Survey},
  year      = {2020},
  journal   = {arXiv preprint arXiv:2006.12057}
}

@inproceedings{Pumarola21DNerf,
  author    = {Pumarola, Albert and Corona, Enric and Pons-Moll, Gerard and Moreno-Noguer, Francesc},
  title     = {{D-NeRF}: Neural radiance fields for dynamic scenes},
  booktitle = {Proceedings of the {IEEE/CVF} Conference on Computer Vision and Pattern Recognition},
  year      = {2021},
  pages     = {10318--10327}
}

@inproceedings{Ost21NeuralSceneGraphs,
  author    = {Ost, Julian and Mannan, Fahim and Thuerey, Nils and Knodt, Julian and Heide, Felix},
  title     = {Neural scene graphs for dynamic scenes},
  booktitle = {Proceedings of the {IEEE/CVF} Conference on Computer Vision and Pattern Recognition},
  year      = {2021},
  pages     = {2856--2865}
}

@inproceedings{Tretschk21NonRigidNeural,
  author    = {Tretschk, Edgar and Tewari, Ayush and Golyanik, Vladislav and Zollh{\"{o}}fer, Michael and Lassner, Christoph and Theobalt, Christian},
  title     = {Non-rigid neural radiance fields: Reconstruction and novel view synthesis of a dynamic scene from monocular video},
  booktitle = {Proceedings of the {IEEE/CVF} International Conference on Computer Vision},
  year      = {2021},
  pages     = {12959--12970}
}

@article{Han22VisionTransformerSurvey,
  author    = {Han, Kai and Wang, Yunhe and Chen, Hanting and Chen, Xinghao and Guo, Jianyuan and Liu, Zhenhua and Tang, Yehui and et al.},
  title     = {A survey on vision {Transformer}},
  journal   = {{IEEE} Transactions on Pattern Analysis and Machine Intelligence},
  volume    = {45},
  number    = {1},
  year      = {2022},
  pages     = {87--110}
}

@article{Peng23RWKV,
  author    = {Peng, Bo and Alcaide, Eric and Anthony, Quentin and Albalak, Alon and Arcadinho, Samuel and Biderman, Stella and Cao, Huanqi and et al.},
  title     = {{RWKV}: Reinventing {RNNs} for the {Transformer} era},
  year      = {2023},
  journal   = {arXiv preprint arXiv:2305.13048}
}

@inproceedings{Friedman23LearningTransformerPrograms,
  author    = {Friedman, Dan and Wettig, Alexander and Chen, Danqi},
  title     = {Learning {Transformer} programs},
  booktitle = {Advances in Neural Information Processing Systems},
  volume    = {36},
  year      = {2023},
  pages     = {49044--49067}
}

@inproceedings{Hoover23EnergyTransformer,
  author    = {Hoover, Benjamin and Liang, Yuchen and Pham, Bao and Panda, Rameswar and Strobelt, Hendrik and Chau, Duen Horng and Zaki, Mohammed and Krotov, Dmitry},
  title     = {Energy {Transformer}},
  booktitle = {Advances in Neural Information Processing Systems},
  volume    = {36},
  year      = {2023},
  pages     = {27532--27559}
}

@article{Amatriain23TransformerModelsIntro,
  author    = {Amatriain, Xavier and Sankar, Ananth and Bing, Jie and Bodigutla, Praveen Kumar and Hazen, Timothy J. and Kazi, Michaeel},
  title     = {{Transformer} models: an introduction and catalog},
  year      = {2023},
  journal   = {arXiv preprint arXiv:2302.07730}
}

@inproceedings{Han23FlattenTransformer,
  author    = {Han, Dongchen and Pan, Xuran and Han, Yizeng and Song, Shiji and Huang, Gao},
  title     = {Flatten {Transformer}: Vision {Transformer} using focused linear attention},
  booktitle = {Proceedings of the {IEEE/CVF} International Conference on Computer Vision},
  year      = {2023},
  pages     = {5961--5971}
}

@article{Yao23DualVisionTransformer,
  author    = {Yao, Ting and Li, Yehao and Pan, Yingwei and Wang, Yu and Zhang, Xiao-Ping and Mei, Tao},
  title     = {Dual vision {Transformer}},
  journal   = {{IEEE} Transactions on Pattern Analysis and Machine Intelligence},
  volume    = {45},
  number    = {9},
  year      = {2023},
  pages     = {10870--10882}
}

@inproceedings{Gao23GeneralizedRelationModeling,
  author    = {Gao, Shenyuan and Zhou, Chunluan and Zhang, Jun},
  title     = {Generalized relation modeling for {Transformer} tracking},
  booktitle = {Proceedings of the {IEEE/CVF} Conference on Computer Vision and Pattern Recognition},
  year      = {2023},
  pages     = {18686--18695}
}

@inproceedings{Chen24TextTo3DGaussian,
  author    = {Chen, Zilong and Wang, Feng and Wang, Yikai and Liu, Huaping},
  title     = {Text-to-{3D} using gaussian splatting},
  booktitle = {Proceedings of the {IEEE/CVF} Conference on Computer Vision and Pattern Recognition},
  year      = {2024},
  pages     = {21401--21412}
}

@article{Tang23Dreamgaussian,
  author    = {Tang, Jiaxiang and Ren, Jiawei and Zhou, Hang and Liu, Ziwei and Zeng, Gang},
  title     = {{DreamGaussian}: Generative gaussian splatting for efficient {3D} content creation},
  year      = {2023},
  journal   = {arXiv preprint arXiv:2309.16653}
}

@inproceedings{Yi24GaussianDreamer,
  author    = {Yi, Taoran and Fang, Jiemin and Wang, Junjie and Wu, Guanjun and Xie, Lingxi and Zhang, Xiaopeng and Liu, Wenyu and Tian, Qi and Wang, Xinggang},
  title     = {{GaussianDreamer}: Fast generation from text to {3D} gaussians by bridging {2D} and {3D} diffusion models},
  booktitle = {Proceedings of the {IEEE/CVF} Conference on Computer Vision and Pattern Recognition},
  year      = {2024},
  pages     = {6796--6807}
}

@article{Luo22DualGGAN,
  author    = {Luo, Xiaodong and He, Xiaohai and Chen, Xiang and Qing, Linbo and Zhang, Jin},
  title     = {{DualG-GAN}, a Dual-channel Generator based Generative Adversarial Network for text-to-face synthesis},
  journal   = {Neural Networks},
  volume    = {155},
  year      = {2022},
  pages     = {155--167}
}

@article{Yu22CMOSGAN,
  author    = {Yu, Shikang and Han, Hu and Shan, Shiguang and Chen, Xilin},
  title     = {{CMOS-GAN}: Semi-supervised generative adversarial model for cross-modality face image synthesis},
  journal   = {{IEEE} Transactions on Image Processing},
  volume    = {32},
  year      = {2022},
  pages     = {144--158}
}

@article{Kammoun22GANSurvey,
  author    = {Kammoun, Amina and Slama, Rim and Tabia, Hedi and Ouni, Tarek and Abid, Mohmed},
  title     = {Generative adversarial networks for face generation: A survey},
  journal   = {{ACM} Computing Surveys},
  volume    = {55},
  number    = {5},
  year      = {2022},
  articleno = {100}
}

@inproceedings{Pesavento24ANIM,
  author    = {Pesavento, Marco and Xu, Yuanlu and Sarafianos, Nikolaos and Maier, Robert and Wang, Ziyan and Yao, Chun-Han and Volino, Marco and Boyer, Edmond and Hilton, Adrian and Tung, Tony},
  title     = {{ANIM}: Accurate neural implicit model for human reconstruction from a single {RGB-D} image},
  booktitle = {Proceedings of the {IEEE/CVF} Conference on Computer Vision and Pattern Recognition},
  year      = {2024},
  pages     = {5448--5458}
}

@article{Mildenhall21Nerf,
  author    = {Mildenhall, Ben and Srinivasan, Pratul P. and Tancik, Matthew and Barron, Jonathan T. and Ramamoorthi, Ravi and Ng, Ren},
  title     = {{NeRF}: Representing scenes as neural radiance fields for view synthesis},
  journal   = {Communications of the {ACM}},
  volume    = {65},
  number    = {1},
  year      = {2021},
  pages     = {99--106}
}

@inproceedings{Athar22Rignerf,
  author    = {Athar, ShahRukh and Xu, Zexiang and Sunkavalli, Kalyan and Shechtman, Eli and Shu, Zhixin},
  title     = {{RigNeRF}: Fully controllable neural {3D} portraits},
  booktitle = {Proceedings of the {IEEE/CVF} Conference on Computer Vision and Pattern Recognition},
  year      = {2022},
  pages     = {20364--20373}
}

@inproceedings{Grassal22NeuralHead,
  author    = {Grassal, Philip-William and Prinzler, Malte and Leistner, Titus and Rother, Carsten and Nie{\ss}ner, Matthias and Thies, Justus},
  title     = {Neural head avatars from monocular {RGB} videos},
  booktitle = {Proceedings of the {IEEE/CVF} Conference on Computer Vision and Pattern Recognition},
  year      = {2022},
  pages     = {18653--18664}
}

@inproceedings{Bergman22GenerativeNeural,
  author    = {Bergman, Alexander and Kellnhofer, Petr and Wang, Yifan and Chan, Eric and Lindell, David and Wetzstein, Gordon},
  title     = {Generative neural articulated radiance fields},
  booktitle = {Advances in Neural Information Processing Systems 35},
  year      = {2022},
  pages     = {19900--19916}
}

@inproceedings{Jiang22Selfrecon,
  author    = {Jiang, Boyi and Hong, Yang and Bao, Hujun and Zhang, Juyong},
  title     = {{SelfRecon}: Self reconstruction your digital avatar from monocular video},
  booktitle = {Proceedings of the {IEEE/CVF} Conference on Computer Vision and Pattern Recognition},
  year      = {2022},
  pages     = {5605--5615}
}

@inproceedings{Cui24AlethNerf,
  author    = {Cui, Ziteng and Gu, Lin and Sun, Xiao and Ma, Xianzheng and Qiao, Yu and Harada, Tatsuya},
  title     = {{Aleth-NeRF}: Illumination adaptive {NeRF} with concealing field assumption},
  booktitle = {Proceedings of the {AAAI} Conference on Artificial Intelligence},
  volume    = {38},
  number    = {2},
  year      = {2024},
  pages     = {1435--1444}
}

@inproceedings{Zhan24KFDNeRF,
  author    = {Zhan, Yifan and Li, Zhuoxiao and Niu, Muyao and Zhong, Zhihang and Nobuhara, Shohei and Nishino, Ko and Zheng, Yinqiang},
  title     = {{KFD-NeRF}: Rethinking Dynamic {NeRF} with Kalman Filter},
  booktitle = {European Conference on Computer Vision},
  publisher = {Springer Nature Switzerland},
  address   = {Cham},
  year      = {2024},
  pages     = {1--18}
}

@inproceedings{Zheng22ImAvatar,
  author    = {Zheng, Yufeng and Fern{\'{a}}ndez Abrevaya, Victoria and B{\"{u}}hler, Marcel C. and Chen, Xu and Black, Michael J. and Hilliges, Otmar},
  title     = {{IMAvatar}: Implicit morphable head avatars from videos},
  booktitle = {Proceedings of the {IEEE/CVF} Conference on Computer Vision and Pattern Recognition},
  year      = {2022},
  pages     = {13545--13555}
}

@article{Kerbl23GaussianSplatting,
  author    = {Kerbl, Bernhard and Kopanas, Georgios and Leimk{\"{u}}hler, Thomas and Drettakis, George},
  title     = {{3D} gaussian splatting for real-time radiance field rendering},
  journal   = {{ACM} Trans. Graph.},
  volume    = {42},
  number    = {4},
  year      = {2023},
  articleno = {139},
  numpages  = {14}
}

@article{Fei24GaussianSplattingSurvey,
  author    = {Fei, Ben and Xu, Jingyi and Zhang, Rui and Zhou, Qingyuan and Yang, Weidong and He, Ying},
  title     = {{3D} gaussian splatting as new era: A survey},
  journal   = {{IEEE} Transactions on Visualization and Computer Graphics},
  year      = {2024},
  note      = {To appear}
}

@article{Ali25CompressionGaussian,
  author    = {Ali, Muhammad Salman and Zhang, Chaoning and Cagnazzo, Marco and Valenzise, Giuseppe and Tartaglione, Enzo and Bae, Sung-Ho},
  title     = {Compression in {3D} Gaussian Splatting: A Survey of Methods, Trends, and Future Directions},
  year      = {2025},
  journal   = {arXiv preprint arXiv:2502.19457}
}

@article{Bao25_3DGaussianSplattingSurvey,
  author    = {Bao, Yanqi and Ding, Tianyu and Huo, Jing and Liu, Yaoli and Li, Yuxin and Li, Wenbin and Gao, Yang and Luo, Jiebo},
  title     = {{3D} gaussian splatting: Survey, technologies, challenges, and opportunities},
  journal   = {{IEEE} Transactions on Circuits and Systems for Video Technology},
  year      = {2025},
  note      = {To appear}
}

@inproceedings{Zielonka23InstantVolumetric,
  author    = {Zielonka, Wojciech and Bolkart, Timo and Thies, Justus},
  title     = {Instant volumetric head avatars},
  booktitle = {Proceedings of the {IEEE/CVF} Conference on Computer Vision and Pattern Recognition},
  year      = {2023},
  pages     = {4574--4584}
}

@inproceedings{Siarohin19FirstOrderMotion,
  author    = {Siarohin, Aliaksandr and Lathuili{\`{e}}re, St{\'{e}}phane and Tulyakov, Sergey and Ricci, Elisa and Sebe, Nicu},
  title     = {First order motion model for image animation},
  booktitle = {Advances in Neural Information Processing Systems 32},
  year      = {2019}
}

@inproceedings{Gafni21DynamicNeural,
  author    = {Gafni, Guy and Thies, Justus and Zollhofer, Michael and Nie{\ss}ner, Matthias},
  title     = {Dynamic neural radiance fields for monocular {4D} facial avatar reconstruction},
  booktitle = {Proceedings of the {IEEE/CVF} Conference on Computer Vision and Pattern Recognition},
  year      = {2021},
  pages     = {8649--8658}
}

@inproceedings{Hong22Headnerf,
  author    = {Hong, Yang and Peng, Bo and Xiao, Haiyao and Liu, Ligang and Zhang, Juyong},
  title     = {{HeadNeRF}: A real-time {NeRF}-based parametric head model},
  booktitle = {Proceedings of the {IEEE/CVF} Conference on Computer Vision and Pattern Recognition},
  year      = {2022},
  pages     = {20374--20384}
}

@inproceedings{Zheng23Pointavatar,
  author    = {Zheng, Yufeng and Wang, Yifan and Wetzstein, Gordon and Black, Michael J. and Hilliges, Otmar},
  title     = {{PointAvatar}: Deformable point-based head avatars from videos},
  booktitle = {Proceedings of the {IEEE/CVF} Conference on Computer Vision and Pattern Recognition},
  year      = {2023},
  pages     = {21057--21067}
}

@article{Paszke19Pytorch,
  author    = {Paszke, Adam and Gross, Sam and Massa, Francisco and Lerer, Adam and Bradbury, James and Chanan, Gregory and Killeen, Trevor and Lin, Zeming and Gimelshein, Natalia and Antiga, Luca and Desmaison, Alban and Kopf, Andreas and Yang, Edward and DeVito, Zachary and Raison, Martin and Tejani, Alykhan and Chilamkurthy, Sasank and Steiner, Benoit and Fang, Lu and Bai, Junjie and Chintala, Soumith},
  title     = {{PyTorch}: An imperative style, high-performance deep learning library},
  year      = {2019},
  journal   = {arXiv preprint arXiv:1912.01703}
}

@inproceedings{Ahmed24LinguisticIntelligence,
  author    = {Ahmed, Tasnim and Piovesan, Nicola and De Domenico, Antonio and Choudhury, Salimur},
  title     = {Linguistic intelligence in large language models for telecommunications},
  booktitle = {2024 {IEEE} International Conference on Communications Workshops ({ICC} Workshops)},
  publisher = {IEEE},
  year      = {2024},
  pages     = {1237--1243}
}

@article{Yu23SocialVision,
  author    = {Yu, Hui and Wang, Yutong and Tian, Yonglin and Zhang, Hui and Zheng, Wenbo and Wang, Fei-Yue},
  title     = {Social vision for intelligent vehicles: From computer vision to foundation vision},
  journal   = {{IEEE} Transactions on Intelligent Vehicles},
  volume    = {8},
  number    = {11},
  year      = {2023},
  pages     = {4474--4476}
}

@article{Chen20CitywideTraffic,
  author    = {Chen, Cen and Li, Kenli and Teo, Sin G. and Zou, Xiaofeng and Li, Keqin and Zeng, Zeng},
  title     = {Citywide traffic flow prediction based on multiple gated spatio-temporal convolutional neural networks},
  journal   = {{ACM} Transactions on Knowledge Discovery from Data ({TKDD})},
  volume    = {14},
  number    = {4},
  year      = {2020},
  articleno = {41},
  numpages  = {23}
}

@incollection{Blanz23MorphableModel,
  author    = {Blanz, Volker and Vetter, Thomas},
  title     = {A morphable model for the synthesis of {3D} faces},
  booktitle = {Seminal Graphics Papers: Pushing the Boundaries, Volume 2},
  year      = {2023},
  pages     = {157--164}
}

@article{Thies19DeferredNeural,
  author    = {Thies, Justus and Zollh{\"o}fer, Michael and Nie{\ss}ner, Matthias},
  title     = {Deferred neural rendering: Image synthesis using neural textures},
  journal   = {{ACM} Transactions on Graphics ({TOG})},
  volume    = {38},
  number    = {4},
  year      = {2019},
  articleno = {67},
  numpages  = {12}
}

@inproceedings{Buehler21Varitex,
  author    = {B{\"{u}}hler, Marcel C. and Meka, Abhimitra and Li, Gengyan and Beeler, Thabo and Hilliges, Otmar},
  title     = {{Varitex}: Variational neural face textures},
  booktitle = {Proceedings of the {IEEE/CVF} International Conference on Computer Vision},
  year      = {2021},
  pages     = {13890--13899}
}

@inproceedings{Xiang24Flashavatar,
  author    = {Xiang, Jun and Gao, Xuan and Guo, Yudong and Zhang, Juyong},
  title     = {{FlashAvatar}: High-fidelity head avatar with efficient gaussian embedding},
  booktitle = {Proceedings of the {IEEE/CVF} Conference on Computer Vision and Pattern Recognition},
  year      = {2024},
  pages     = {1802--1812}
}

@inproceedings{Wu24_4DGaussianSplatting,
  author    = {Wu, Guanjun and Yi, Taoran and Fang, Jiemin and Xie, Lingxi and Zhang, Xiaopeng and Wei, Wei and Liu, Wenyu and Tian, Qi and Wang, Xinggang},
  title     = {{4D} gaussian splatting for real-time dynamic scene rendering},
  booktitle = {Proceedings of the {IEEE/CVF} Conference on Computer Vision and Pattern Recognition},
  year      = {2024},
  pages     = {20310--20320}
}

@article{Yu12PerceptionDriven,
  author    = {Yu, Hui and Garrod, Oliver GB and Schyns, Philippe G.},
  title     = {Perception-driven facial expression synthesis},
  journal   = {Computers \& Graphics},
  volume    = {36},
  number    = {3},
  year      = {2012},
  pages     = {152--162}
}

@inproceedings{Vaswani17Attention,
  author    = {Vaswani, Ashish and Shazeer, Noam and Parmar, Niki and Uszkoreit, Jakob and Jones, Llion and Gomez, Aidan N. and Kaiser, {\L}ukasz and Polosukhin, Illia},
  title     = {Attention is all you need},
  booktitle = {Advances in Neural Information Processing Systems 30},
  year      = {2017}
}

@inproceedings{Fan22Faceformer,
  author    = {Fan, Yingruo and Lin, Zhaojiang and Saito, Jun and Wang, Wenping and Komura, Taku},
  title     = {{Faceformer}: Speech-driven {3D} facial animation with transformers},
  booktitle = {Proceedings of the {IEEE/CVF} Conference on Computer Vision and Pattern Recognition},
  year      = {2022},
  pages     = {18770--18780}
}

@inproceedings{Yang24DeformableGaussians,
  author    = {Yang, Ziyi and Gao, Xinyu and Zhou, Wen and Jiao, Shaohui and Zhang, Yuqing and Jin, Xiaogang},
  title     = {Deformable {3D} gaussians for high-fidelity monocular dynamic scene reconstruction},
  booktitle = {Proceedings of the {IEEE/CVF} Conference on Computer Vision and Pattern Recognition},
  year      = {2024},
  pages     = {20331--20341}
}

@inproceedings{Chibane21StereoRadiance,
  author    = {Chibane, Julian and Bansal, Aayush and Lazova, Verica and Pons-Moll, Gerard},
  title     = {Stereo radiance fields ({SRF}): Learning view synthesis for sparse views of novel scenes},
  booktitle = {Proceedings of the {IEEE/CVF} Conference on Computer Vision and Pattern Recognition},
  year      = {2021},
  pages     = {7911--7920}
}

@article{Kingma14Adam,
  author    = {Kingma, Diederik P. and Ba, Jimmy},
  title     = {Adam: A method for stochastic optimization},
  year      = {2014},
  journal   = {arXiv preprint arXiv:1412.6980}
}

@inproceedings{Kwak20CafeGAN,
  author    = {Kwak, Jeong-gi and Han, David K. and Ko, Hanseok},
  title     = {{CAFE-GAN}: Arbitrary face attribute editing with complementary attention feature},
  booktitle = {Computer Vision--{ECCV} 2020: 16th European Conference, Glasgow, {UK}, August 23--28, 2020, Proceedings, Part {XIV}},
  series    = {Lecture Notes in Computer Science},
  volume    = {12359},
  publisher = {Springer International Publishing},
  year      = {2020},
  pages     = {524--540}
}

@article{Chen22Transformer3DFace,
  author    = {Chen, Zhuo and Wang, Yuesong and Guan, Tao and Xu, Luoyuan and Liu, Wenkai},
  title     = {Transformer-based {3D} face reconstruction with end-to-end shape-preserved domain transfer},
  journal   = {{IEEE} Transactions on Circuits and Systems for Video Technology},
  volume    = {32},
  number    = {12},
  year      = {2022},
  pages     = {8383--8393}
}

@article{Wang23ChatGPTDAO,
  author    = {Wang, Fei-Yue and Miao, Qinghai and Li, Xuan and Wang, Xingxia and Lin, Yilun},
  title     = {What does {ChatGPT} say: The {DAO} from algorithmic intelligence to linguistic intelligence},
  journal   = {{IEEE/CAA} Journal of Automatica Sinica},
  volume    = {10},
  number    = {3},
  year      = {2023},
  pages     = {575--579}
}

@inproceedings{zhang2022transnerf,
  author    = {Zhang, Qi and Yang, Mingchuan and Zou, Hang and Liu, Qiaoqiao},
  title     = {{TransNeRF}: Multi-View Optimization for General Neural Radiance Fields Across Scenes},
  booktitle = {Proceedings of the International Conference on Virtual Reality and Human-Computer Interaction and Artificial Intelligence ({VRHCIAI})},
  year      = {2022},
  pages     = {111--116},
  publisher = {IEEE}
}

@inproceedings{Deng22Gram,
  author    = {Deng, Yu and Yang, Jiaolong and Xiang, Jianfeng and Tong, Xin},
  title     = {{GRAM}: Generative radiance manifolds for {3D}-aware image generation},
  booktitle = {Proceedings of the {IEEE/CVF} Conference on Computer Vision and Pattern Recognition},
  year      = {2022},
  pages     = {10673--10683}
}

@inproceedings{Chan22EfficientGeometry,
  author    = {Chan, Eric R. and Lin, Connor Z. and Chan, Matthew A. and Nagano, Koki and Pan, Boxiao and De Mello, Shalini and Gallo, Orazio and et al.},
  title     = {Efficient geometry-aware {3D} generative adversarial networks},
  booktitle = {Proceedings of the {IEEE/CVF} Conference on Computer Vision and Pattern Recognition},
  year      = {2022},
  pages     = {16123--16133}
}

@inproceedings{Choi18Stargan,
  author    = {Choi, Yunjey and Choi, Minje and Kim, Munyoung and Ha, Jung-Woo and Kim, Sunghun and Choo, Jaegul},
  title     = {{StarGAN}: Unified generative adversarial networks for multi-domain image-to-image translation},
  booktitle = {Proceedings of the {IEEE} Conference on Computer Vision and Pattern Recognition},
  year      = {2018},
  pages     = {8789--8797}
}

@article{Wang04ImageQuality,
  author    = {Wang, Zhou and Bovik, Alan C. and Sheikh, Hamid R. and Simoncelli, Eero P.},
  title     = {Image quality assessment: from error visibility to structural similarity},
  journal   = {{IEEE} Transactions on Image Processing},
  volume    = {13},
  number    = {4},
  year      = {2004},
  pages     = {600--612}
}

@inproceedings{Zhang18UnreasonableEffectiveness,
  author    = {Zhang, Richard and Isola, Phillip and Efros, Alexei A. and Shechtman, Eli and Wang, Oliver},
  title     = {The unreasonable effectiveness of deep features as a perceptual metric},
  booktitle = {Proceedings of the {IEEE} Conference on Computer Vision and Pattern Recognition},
  year      = {2018},
  pages     = {586--595}
}

\end{document}